%% file: neurips_2026.tex
\documentclass{article}

\PassOptionsToPackage{numbers, compress}{natbib}
\usepackage[main, final]{neurips_2026}

\usepackage[utf8]{inputenc} 
\usepackage[T1]{fontenc}    
\usepackage{hyperref}       
\usepackage{url}            
\usepackage{booktabs}       
\usepackage{amsfonts}       
\usepackage{nicefrac}       
\usepackage{microtype}      
\usepackage{xcolor}         
\usepackage{multirow}
\usepackage{graphicx}
\usepackage{amsmath}
\usepackage{tcolorbox}
\usepackage{pgfplots}
\pgfplotsset{compat=1.18}
\usepgfplotslibrary{groupplots,fillbetween}
\input{figures/cons_style}
\DeclareMathOperator*{\argmax}{arg\,max}

\title{Consilience for Verifier-Free Test-Time Scaling}

\author{%
  Lecheng Kong, Like Hui, Haitao Mao, Jun Huan\\
  AWS AI Labs\\
  \texttt{\{jkong,likehui,maoht,lukehuan\}@amazon.com}
}

\begin{document}

\maketitle

\begin{center}
\vspace{-1.5em}
\small\textbf{Project page w/ interactive demo:}
\href{https://consilience-tts.github.io/}{{\fontencoding{T1}\fontfamily{zi4}\selectfont
https://consilience-tts.github.io/}}
\end{center}

\begin{abstract}
  Test-time scaling often uses an external verifier, such as compilers and test cases in coding or trained value functions in robotics applications, to obtain high-quality rollouts. Verifier-free test-time scaling (or VF-TTS) is gaining extensive attention as a mechanism to enhance Large Language Model (LLM) reasoning, primarily because we do not have access to such high-quality verifiers in many real-world applications. Among existing VF-TTS methods, confidence-based VF-TTS methods, which compute and rank rollouts solely by confidence, are particularly promising. Such methods introduce near-zero overhead for sample evaluation and require minimal access to internal model states, making the methods highly flexible across models and tasks. 
  In this paper, we demonstrate a critical limitation of existing confidence-based VF-TTS methods by showing that such methods catastrophically break down on complex tasks. We observe a very interesting phenomenon: uniformly high confidence frequently indicates a failure to explore, favoring confidently wrong answers. To address this, our core insight is that robust cognitive search requires a specific confidence trajectory pattern: such methods perform exploratory branching at the beginning, as manifested by low initial confidence, and converge to a high final confidence solution. To implement this insight, we introduce consilience, a novel selection framework that explicitly evaluates the temporal asymmetry of confidence in reasoning. We operationalize this via a combinatorial metric that actively penalizes high initial confidence while strictly demanding final certainty. Extensive experiments covering both graduate-level mathematics problems and free-form code generation demonstrate that consilience effectively outperforms existing baselines, validating our novel perspective on completion confidence.
\end{abstract}

\section{Introduction}
\input{tex/intro}
\section{Related Work}
\input{tex/related_work.tex}
\section{Method}
\input{tex/consilience}
\input{tex/method}

\section{Experiment}
\input{tex/exp}
\section{Conclusion, Limitations, and Future Work}\label{sec:conclusion}
We introduced consilience, a verifier-free test-time scaling framework that replaces confidence maximization with a temporal signal: low confidence while exploring, high confidence at convergence. It neutralizes confidently wrong selections and achieves significant margins over existing baselines, particularly on open-ended code generation, where exact-match majority voting is not applicable. More broadly, consilience instantiates a general test-time-scaling principle, that exploration should be valued rather than penalized. Our score is one efficient, zero-overhead instantiation of it, and a fixed heuristic rather than a universal rule: some settings may admit multiple valid confidence trajectories. Our agentic integration is also a coarse, keyword-triggered harness, which limits efficiency. Future directions include learned or judge-based selectors that evaluate the exploratory diversity of the reasoning itself; using consilience to select among multiple shallow trees under a fixed budget, mitigating the outsized influence of early sampling in tree search; smarter agentic harnesses; and, since consilience is an entirely internal measure of reasoning quality, an intrinsic reward for verifier-free agentic reinforcement learning.
\section*{Acknowledgments}
The authors would like to thank Narayanan Sadagopan and Vijay Lingam for helpful feedback and discussion.

\bibliography{neurips2026_conference}
\bibliographystyle{plainnat}

\newpage
\appendix
\section*{Appendix}
\section{Dataset Details}
\input{tex/app_dataset.tex}
\section{Model Configuration}
\input{tex/app_model.tex}

\section{Voting-based Results}
\input{tex/app_majority_vote.tex}
\section{Statistical Significance of the Confidence Trajectory}
\input{tex/app_stat.tex}
\section{Practical Budget and Extended Baselines on Common-Sense QA}
\input{tex/app_nq.tex}
\section{More Related Work}
\input{tex/app_more_rel}
\section{Qualitative Analysis}
\input{tex/app_qualitative}
\section{Prompt for Generation}
\input{tex/app_prompts.tex}
\section{Agentic Harness for Consilience}
\input{tex/app_agentic.tex}

\end{document}

%% file: figures/cons_style.tex
\definecolor{csCons}{HTML}{2A78D6}
\definecolor{csConsDark}{HTML}{184F95}
\definecolor{csOver}{HTML}{EB6834}
\definecolor{csLast}{HTML}{1BAF7A}
\definecolor{csBase}{HTML}{898781}
\definecolor{csNeg}{HTML}{E34948}
\definecolor{csMid}{HTML}{F0EFEC}
\definecolor{csInk}{HTML}{0B0B0B}
\definecolor{csInkTwo}{HTML}{52514E}
\definecolor{csMuted}{HTML}{898781}
\definecolor{csGrid}{HTML}{E1E0D9}
\pgfplotsset{
  cs panel/.style={
    scale only axis,
    tick label style={font=\scriptsize, text=csInk},
    label style={font=\scriptsize, text=csInk},
    title style={font=\scriptsize\bfseries, text=csInk, yshift=-6pt},
    xlabel style={yshift=5pt},
    ylabel style={yshift=-5pt},
    grid=major,
    major grid style={draw=csGrid, line width=0.4pt},
    axis line style={draw=csMuted!70, line width=0.4pt},
    tick style={draw=csMuted!70},
    major tick length=2pt,
    yticklabel style={/pgf/number format/fixed},
    scaled y ticks=false,
    clip=false,
  },
  cs legend/.style={
    legend style={draw=none, fill=none, font=\scriptsize, text=csInk,
      legend columns=-1, /tikz/every even column/.append style={column sep=8pt}},
    legend cell align=left,
  },
  cs line/.style={line width=0.9pt, line join=round, line cap=round, mark=none},
  cs dot/.style={mark=*, mark size=1.15pt, mark options={draw=white, line width=0.25pt, fill=#1}},
}
\tikzset{
  cs note/.style={font=\scriptsize, text=csInk, inner sep=1pt},
}

%% file: tex/intro.tex
Reasoning models have demonstrated remarkable capabilities across diverse domains. In reasoning-heavy scenarios where accuracy is of utmost importance, parallel test-time scaling (TTS) has emerged as a highly effective method for improving model performance--generating multiple reasoning paths during inference and selecting the optimal output~\citep{snell2024scalingllmtesttimecompute,yao2023treethoughtsdeliberateproblem,Besta_2024}. However, the prevailing mechanisms for selecting the optimal output from these parallel traces rely heavily on external verifiers~\citep{snell2024scalingllmtesttimecompute,setlur2025scalingtesttimecomputeverification,zheng2025opencodeinterpreterintegratingcodegeneration,le2022coderlmasteringcodegeneration} or models~\citep{lightman2023letsverifystepstep,yao2023treethoughtsdeliberateproblem}. To scale test-time computation universally in unverifiable scenarios like free-form code generation and intermediate agentic steps, the research community recently started to explore selection mechanisms that rely entirely on the model's self-contained internal signals, such as token-level log-probabilities, to obtain high-quality rollouts. 

Existing log-probability-based methods compute and select maximally confident completions~\citep{kang2025scalablebestofnselectionlarge,fu2025deepthinkconfidence}. The key insight behind these approaches is: higher confidence implies a higher chance of correctness of the rollout. In Figure~\ref{fig:init_hist} (Left), we plot the histogram of the average sequence confidence, and found that the correct answer distribution has a higher mean than the incorrect ones, aligning with the insight aforementioned. However, we discovered that this simple maximization catastrophically breaks down on complex tasks (i.e., model achieving lower correct rate on these problems). It overlooks a critical cognitive factor necessary for solving complex problems: process diversity. To study this insight, in Figure~\ref{fig:init_hist} (Middle), we plot the same histogram except we only include questions where the model achieves $<20\%$ correct rate. In these difficult tasks, the pattern is reversed, and the distributions become inseparable. This result indicates that confidence maximization is useful in solving problems where a model has good answers to such problems, but confidence maximization is less effective where the model has to explore several paths to identify the right solution. 

In this paper, we have investigated the problem of effective TTS for hard problems through the lens of consilience—the principle that a \textbf{conclusion} is reliable only if the model \textbf{knows diverse paths} to solve the problem and all of the paths converge to the same solution~\citep{wang2023selfconsistencyimproveschainthought,chen2023universalselfconsistencylargelanguage}. In confidence maximization, however, the confidence of both the answer and reasoning are maximized. That is, the selected completions not only converge at the answer, but also converge early in the reasoning phase. Such selections blindly commit to a single line of thought without the model internally acknowledging alternatives. Hence, "confidently wrong" selection happens because confidence maximization forces the system to focus solely on \textbf{quickly reaching a conclusion} or convergence while suppressing the importance of \textbf{knowing diverse paths}, breaking the principle of consilience. Fortunately, this observation naturally shed lights on how to enforce the principle: if high confidence represents convergence, we can use low confidence to identify essential process diversity. If a reasoning process shows a robust understanding of a problem, its awareness of alternative solutions forces the probability mass to split, naturally manifesting as a drop in early confidence. A consilient trace should therefore exhibit a rising trajectory: it begins with an exploratory phase characterized by lower confidence as the model recognizes various valid approaches, which then converges into a highly certain conclusion.

To implement this structural signature, we introduce the metric of consilience, a novel, verifier-free selection framework that explicitly quantifies this temporal confidence asymmetry of effective reasoning. Rather than uniformly maximizing sequence-level confidence, our approach evaluates the trajectory via a combined metric that penalizes high initial confidence while still encouraging high final confidence. Using our metric, we can re-plot the histogram of our metric on the hard problems in Figure~\ref{fig:init_hist} (right), and show that consilience can successfully recover the pattern to distinguish correct and incorrect responses. Extensive experiments across complex mathematics (HMMT), graduate-level knowledge (GPQA), and open-ended code generation (LiveCodeBench-V6, agentic Swe-Bench) demonstrate the efficacy of this approach. Most notably, consilience unlocks significant margins in free-form coding domains where voting-based technique is inapplicable; on LiveCodeBench, our metric elevates the GPT-OSS-120B model to an accuracy of 69.7\%, substantially outperforming Pass@1 (65.7\%). Similar systematic gains are observed in other reasoning-heavy datasets.
\input{block/init_hist.tex}

%% file: block/init_hist.tex
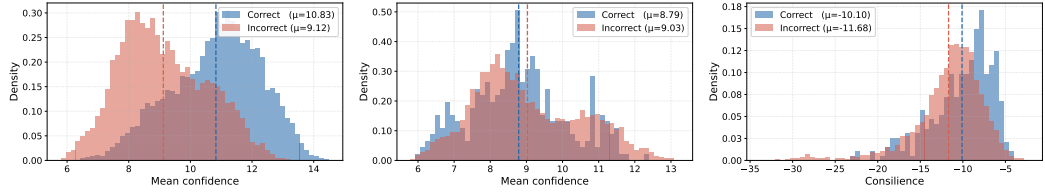
\begin{figure}
    \centering
    \input{figures/fig_init_hist}
    \caption{Left: Histogram of completion confidence on LiveCodeBench-V6 using GPT-OSS-120B model. Middle: Same setting as Left but only on hard problems with < 20\% correct rate, where incorrect
answers exhibit a higher mean confidence ($\mu$ = 9.03) than correct ones ($\mu$ = 8.79). Right: Consilience histogram on the hard problems. Confidence is effective on a macro level, but breaks on hard problems. However, our approach, consilience, can maintain the effectiveness.}
    \label{fig:init_hist}
\end{figure}

%% file: figures/fig_init_hist.tex
\begin{tikzpicture}
\begin{groupplot}[
  group style={group size=3 by 1, horizontal sep=0.95cm},
  cs panel, cs legend,
  width=3.6cm, height=1.27cm,
  ymin=0,
  grid=none, ymajorgrids,
]
\nextgroupplot[
  xmin=5.79, xmax=14.51, ymax=0.40, ytick={0,0.1,0.2,0.3},
  xlabel={Mean confidence (all problems)}, ylabel={Density},
  legend style={at={(axis description cs:1.764,1.03)}, anchor=south},
]
\addlegendimage{area legend, draw=csCons, fill=csCons, fill opacity=0.25}\addlegendentry{Correct}
\addlegendimage{area legend, draw=csOver, fill=csOver, fill opacity=0.25}\addlegendentry{Incorrect}
\addlegendimage{draw=csInkTwo, line width=0.8pt}\addlegendentry{Class mean $\mu$}
\addplot[const plot mark left, draw=csOver, fill=csOver, fill opacity=0.15, line width=0.6pt, forget plot]
  table[x=x, y=incorrect] {figures/data/hist_all_mean.dat} \closedcycle;
\addplot[const plot mark left, draw=csCons, fill=csCons, fill opacity=0.15, line width=0.6pt, forget plot]
  table[x=x, y=correct] {figures/data/hist_all_mean.dat} \closedcycle;
\draw[csOver, line width=0.8pt] (axis cs:9.122,0) -- (axis cs:9.122,0.36);
\draw[csCons, line width=0.8pt] (axis cs:10.828,0) -- (axis cs:10.828,0.36);
\node[cs note, anchor=east] at (axis cs:9.05,0.355) {9.12};
\node[cs note, anchor=west] at (axis cs:10.90,0.355) {10.83};

\nextgroupplot[
  xmin=5.79, xmax=13.23, ymax=0.66, ytick={0,0.2,0.4,0.6},
  xlabel={Mean confidence (hard problems)},
]
\addplot[const plot mark left, draw=csOver, fill=csOver, fill opacity=0.15, line width=0.6pt]
  table[x=x, y=incorrect] {figures/data/hist_hard_mean.dat} \closedcycle;
\addplot[const plot mark left, draw=csCons, fill=csCons, fill opacity=0.15, line width=0.6pt]
  table[x=x, y=correct] {figures/data/hist_hard_mean.dat} \closedcycle;
\draw[csCons, line width=0.8pt] (axis cs:8.786,0) -- (axis cs:8.786,0.60);
\draw[csOver, line width=0.8pt] (axis cs:9.029,0) -- (axis cs:9.029,0.60);
\node[cs note, anchor=east] at (axis cs:8.72,0.59) {8.79};
\node[cs note, anchor=west] at (axis cs:9.10,0.59) {9.03};

\nextgroupplot[
  xmin=-34.44, xmax=-2.23, ymax=0.215, ytick={0,0.05,0.1,0.15,0.2},
  yticklabels={0,0.05,0.10,0.15,0.20},
  xlabel={Consilience (hard problems)},
]
\addplot[const plot mark left, draw=csOver, fill=csOver, fill opacity=0.15, line width=0.6pt]
  table[x=x, y=incorrect] {figures/data/hist_hard_cons.dat} \closedcycle;
\addplot[const plot mark left, draw=csCons, fill=csCons, fill opacity=0.15, line width=0.6pt]
  table[x=x, y=correct] {figures/data/hist_hard_cons.dat} \closedcycle;
\draw[csOver, line width=0.8pt] (axis cs:-11.68,0) -- (axis cs:-11.68,0.195);
\draw[csCons, line width=0.8pt] (axis cs:-10.097,0) -- (axis cs:-10.097,0.195);
\node[cs note, anchor=east] at (axis cs:-11.95,0.192) {$-$11.68};
\node[cs note, anchor=west] at (axis cs:-9.83,0.192) {$-$10.10};
\end{groupplot}
\end{tikzpicture}

%% file: tex/related_work.tex
\textbf{Test-time scaling} (TTS) devotes more test-time computation to achieve better performance. Distinct from sequential TTS signified by thinking models~\citep{openai2024openaio1card}, this paper focuses on parallel TTS. Parallel TTS was initially effective but relied on strict environmental dependencies such as external verifiers~\citep{snell2024scalingllmtesttimecompute,setlur2025scalingtesttimecomputeverification}, causing serious latency and generalizability issues. To expand utility to broader tasks, methods such as Self-Consistency~\citep{wang2023selfconsistencyimproveschainthought} take a majority vote over sampled answers, but this imposes a strict constraint on the answer format (multiple-choice, integers) and does not apply to free-form generation. Another line of work uses the LLM itself to provide feedback—via LLM-based evaluators, process reward models, or self-refinement~\citep{chen2023universalselfconsistencylargelanguage,lightman2023letsverifystepstep,madaan2023selfrefineiterativerefinementselffeedback}—but these introduce significant aggregation overhead. Closest to us, several methods use token-level log-probabilities as a direct proxy for correctness to enable highly efficient, verifier-free selection~\citep{kang2025scalablebestofnselectionlarge,fu2025deepthinkconfidence}. Our work builds on the computational advantages of these internal signals but diverges in usage: rather than assuming confidence should be globally maximized, we show that uniformly high confidence is a symptom of premature convergence, and introduce the consilience trajectory to explicitly reward early-stage exploration. A broader discussion of verifier-based TTS, search, and entropy/confidence in calibration and RL is in Appendix~\ref{app:more_rel}.

\textbf{Prefix and temporal confidence.} A growing body of work—much of it concurrent—studies how confidence \emph{evolves} across reasoning rather than its overall level, and our findings are consistent with this literature. On the ``penalize early confidence'' side, \citet{gai2026understandingmitigatingprematureconfidence} show that premature confidence (early commitment followed by rationalization) predicts flawed reasoning and mitigate it with an RL reward for gradual confidence growth; \citet{wang2026inferencetimeoptimizationconfidence} find correct traces exhibit rising confidence and use this ``confidence gain'' to reweight majority voting; \citet{keramati2026earlytokenconfidencepredictsreasoning} find early-token confidence predicts reasoning quality in multi-agent debate; and related work shows the initial reasoning step disproportionately shapes the outcome~\citep{liao2025lostbeginningreasoning}, that per-step confidence carries a temporal signal~\citep{mao2025temporalizingconfidenceevaluationchainofthought}, that reasoning models are better calibrated in expressing confidence~\citep{yoon2025reasoningmodelsbetterexpress}, and that reasoning can nonetheless inflate confidence even when wrong~\citep{Fu_2026}. Seemingly opposed, \citet{otth2025maximizingprefixconfidencetesttimeefficiently} report that \emph{maximizing} prefix confidence improves mathematical reasoning; we reconcile this by noting the effect is difficulty-dependent—on easy problems, where Pass@1 is already high, high prefix confidence is a reliable signal and consilience is effectively neutral, whereas the initial penalty becomes beneficial on hard problems where uniform confidence marks premature convergence (Section~\ref{sec:robust}, Figure~\ref{fig:conf-diff}). We regard this convergence of independent findings as validation of the underlying insight, and our contribution is distinct in both formulation and applicability. First, rather than only observing the pattern or using it to reweight votes, we operationalize it as an \emph{active selection objective} that penalizes high initial confidence while rewarding final convergence. Second, being training-free and computed from logits alone, consilience applies directly to \emph{free-form generation} where voting is inapplicable (code generation, agentic edits)—a regime none of the above addresses, since voting requires extractable answers and the RL approach requires training. Third, and uniquely, we isolate the reasoning phase from the final-answer phase before scoring, removing the artificially confident summarization tokens that dilute the signal.

%% file: tex/consilience.tex
\subsection{Preliminary: Token-Level Confidence}
To understand and evaluate a generation's confidence, we follow the formulation in previous work~\citep{fu2025deepthinkconfidence,kang2025scalablebestofnselectionlarge} and define token-level confidence. For a given prompt $x$ and a generated sequence $y = (y_1, y_2, \dots, y_L)$, the model outputs a probability distribution over the vocabulary at each time step $t$. Instead of relying solely on the probability of the single decoded token $y_t$, the token confidence is collected over the top-$K$ candidate tokens. The token-level confidence $c_{y,t}$ as the negative mean log-probability of these top-$K$ tokens:
\begin{equation}
    c_{y,t} = -\frac{1}{K} \sum_{j=1}^{K} \log P_\theta(v_{t,j} | y_{<t}, x),
\end{equation}
where $v_{t,j}$ represents the $j$-th most probable token in the vocabulary at step $t$. A centralized distribution causes the value to be high, indicating high confidence, whereas a flat distribution leads to low confidence.

\subsection{Why Confidence is Insufficient: The Long-Tail Trap of Hard Problems} \label{sec:cons}
To scale test-time compute without relying on external verifiers or exact-match consensus, recent methods maximize internal confidence metrics across a set of sampled sequences. These are typically implemented as the average token-level confidence across the generated sequence~\citep{kang2025scalablebestofnselectionlarge,fu2025deepthinkconfidence}:
\begin{equation}
y = \argmax_y \frac{1}{L} \sum_{t=1}^L c_{y,t}. \label{eq:conf}
\end{equation}
These approaches operate on a straightforward principle: maximizing a model's confidence maximizes correctness. This direction is particularly valuable for two primary reasons. \textbf{Flexibility.} It requires only token-level logits and no training, making it readily applicable to open-weight models and closed-weight enterprise LLMs (e.g., Google Gemini) that expose log-probabilities via API. \textbf{Efficiency and Latency.} It introduces near-zero computational overhead for evaluating each completion. All completions can be sampled in parallel, providing better latency than sequential scaling.

To validate the efficacy and limitations of confidence maximization, we analyzed the mean confidence distributions of GPT-OSS-120B completions on the LiveCodeBench-V6 (LCB) dataset, a free-form code generation task. In Figure~\ref{fig:init_hist} left, across the entire dataset, higher average confidence strongly correlates with correctness. This confirms the foundational intuition driving existing confidence-based metrics: on a macro level, confident generations are generally more accurate, and existing methods correctly interpret this global signal.

However, this approach becomes counterproductive on complex tasks. When isolated to "hard" problems (correctness rate $< 20\%$, Figure~\ref{fig:init_hist} middle), the confidence dynamic inverts: incorrect answers exhibit a higher mean confidence than correct ones. Crucially, the incorrect distribution on hard problems is not merely shifted—it becomes highly \textbf{over-dispersed}, forming a heavy "long tail" at the highest confidence percentiles. In free-form generation where exact-match majority voting is inapplicable, verifier-free test-time scaling often relies on argmax selection (picking the single highest-scoring completion from the pool). Because of this right-skewed tail, simply maximizing average confidence guarantees that the algorithm will systematically sample "confidently wrong" hallucinations.

\subsection{Understanding Confidence through Consilience}
\label{sec:anatomy}

To understand such contradictory behavior under different task difficulty, we introduce the concept of consilience. In scientific epistemology, the principle of consilience dictates that a conclusion is profoundly more reliable when it emerges from multiple and diverse lines of inquiry rather than a single, isolated path. Within LLM research, this foundational idea has been widely and successfully adopted—most notably through methods like Self-Consistency~\citep{wang2023selfconsistencyimproveschainthought,chen2023universalselfconsistencylargelanguage}. When we evaluate the failure of global confidence maximization through this lens, its drawback becomes apparent. By rewarding traces that exhibit uniformly high confidence from the very first tokens, these metrics optimize for mere convergence and favor completion converging to an unambiguous single path in the quickest manner. Simultaneously, this strategy suppresses the completions that acknowledge diverse paths. This structurally leads to the selection of completion with an overconfident reasoning process.

To ground this insight in the actual mechanics of an LLM, we examine how confidence is computed. In an autoregressive language model, the confidence of generating the $t$-th token $y_t$ given a prompt $x$ and the preceding generated sequence $y_{<t}$ is coupled with its conditional probability: $P_\theta(y_t | y_{<t}, x)$. Crucially, a token's confidence does not measure the objective correctness of the future sequence; rather, it measures how strictly the past reasoning effort ($y_{<t}$) entails the current token. Because of this, the implications of confidence are phase-dependent:

\textbf{Final Stage} (Large $t$): The intuition behind existing confidence maximization is rooted in the final stages of reasoning. As a model successfully navigates its chosen path, the logical constraints should tighten. If confidence is \textbf{high at the end} of the reasoning trace, it indicates that the preceding, lengthy chain of thought has successfully and systematically converged, and the exhaustive reasoning steps inevitably entail the final conclusion.

\textbf{Initial Stage} (Small $t$): In the early steps of generation, high confidence is still a symptom of convergence. However, it implies that a very short prefix has already collapsed the probability space to a singular path. For complex reasoning tasks, this is highly detrimental; it signifies that the model has failed to recognize the difficulty or nuance of the problem, bypassed necessary exploration, and overfitted to a local, often flawed, solution. This blind commitment leads directly to the "confidently wrong" hallucinations observed in the long tail. Conversely, if a model possesses a robust understanding of a complex problem (one requiring exploration of many possibilities), it will naturally acknowledge that multiple viable routes exist to reach a solution, aligning with the consilience principle. This internal awareness of alternative strategies forces the autoregressive probability mass to split, reflecting a necessary cognitive divergence. Consequently, this essential exploratory phase inherently manifests as \textbf{low initial confidence}.

Therefore, genuine reasoning on complex problems is not a flat line of certainty; it is a dynamic trajectory. It demands a temporal asymmetry: low initial confidence resolving into high final confidence. By dissecting a reasoning trace into its initial and final confidence phases, we can categorize the model's generation into four distinct categories as shown in Table~\ref{tab:cognitive_states}.

Prior research correctly recognized the predictive power of confidence, effectively utilizing it to filter out trajectories that fail to reach logical convergence (Categories 2 and 4). However, their application of this signal remains coarse. By relying on simple maximization, existing methods inadvertently optimize for the Category 1 while actively ignoring—or mathematically penalizing—the essential initial uncertainty required for the Low-High Category 3, representing consilience. This structural blind spot is why existing mechanisms systematically underperform on complex reasoning tasks.

In the following sections, we formalize how to implement consilience as a selection metric designed specifically to target and identify generations within this optimal consilient Category.

\input{block/quadrant.tex}

%% file: block/quadrant.tex
\begin{table}[tbp]
  \centering
  \caption{The Four Categories of LLM Reasoning Trajectories}
  \label{tab:cognitive_states}
  \resizebox{0.99\textwidth}{!}{
  \renewcommand{\arraystretch}{1.4} 
  \begin{tabular}{p{2.8cm} p{6cm} p{6cm}}
    \toprule
    & \textbf{High Final Confidence} & \textbf{Low Final Confidence} \\
    \midrule
    \textbf{High Initial \newline Confidence} & 
    \textbf{Premature Convergence (Category 1):} The model immediately collapses onto a singular path. Often occurs on trivially easy queries or when the model overfits to a flawed premise, leading to confidently wrong answers. & 
    \textbf{Degeneration (Category 2):} The model confidently begins a path but encounters logical contradictions or out-of-distribution states later in the sequence, causing the reasoning to break down. \\
    \addlinespace 
    \textbf{Low Initial \newline Confidence} & 
    \textbf{Consilience (Category 3):} The model actively sees diverse paths (low initial confidence) before logically converging on a well-supported, highly certain conclusion. & 
    \textbf{Total Confusion (Category 4):} The prompt is too difficult for the model. It fails to explore meaningful paths initially and fails to reach a logical consensus at the end. \\
    \bottomrule
  \end{tabular}}
\end{table}

%% file: tex/method.tex
\subsection{The Consilience Score Formulation}\label{sec:cons_score}
With the token-level metric $c_{y,t}$ defined, we quantify the consilience trajectory for a generated sequence $y$ of length $L$. From Section~\ref{sec:cons}, a correct, well-explored reasoning path should exhibit low initial confidence (reflecting active path exploration) that eventually converges into high final confidence (reflecting logical consensus). To capture this principle, we use a boundary window of size $W$ to capture the initial and final confidence:
\begin{equation}C_{initial} = \frac{1}{W} \sum_{t=P+1}^{W+P} c_t, \quad C_{final} = \frac{1}{W} \sum_{t=L - W + 1}^{L} c_t\end{equation}
The initial and final confidence is quantified by the average token confidence of the first and last $W$ tokens. Notice that we also skip the first $P$ tokens, because the initial tokens will have very similar confidence (e.g., the first token will have the same confidence given the same prompt), which will dilute the confidence signal.
Finally, we define the Consilience Score ($S$) for a given reasoning trace to explicitly reward high final confidence while penalizing high initial confidence.
\begin{equation}S = C_{final} - \alpha C_{initial},\label{eq:cons}\end{equation}
where $\alpha \ge 0$ is a balancing factor.
During test-time scaling, we sample multiple completions for a given prompt, compute $S$ for each sequence, and select the completion that maximizes $S$. By minimizing initial confidence and maximizing final confidence, this objective actively selects for the diverse exploration and robust convergence that characterizes true consilience. We plot the consilience histogram on the same hard problems set in Figure~\ref{fig:init_hist}, the consilience mean of correct responses is significantly higher than that of incorrect ones. More importantly, the incorrect distribution does not exhibit an overly dispersed long-tail distribution, making consilience maximization a stable metric for free-form generation problems.

\subsection{Reasoning-Phase Isolation}
A critical nuance in applying confidence metrics to models trained with Chain-of-Thought (CoT) or explicit \texttt{<think>} tokens lies in the structural composition of the generated sequence. Autoregressive outputs under this paradigm typically consist of two distinct semantic phases: the active reasoning phase ($Y_{reason}$) and the final answer formulation ($Y_{answer}$). Specifically, the tokens within $Y_{answer}$ do not reflect active problem-solving or cognitive search. These tokens mostly serve to summarize or mimic the logical conclusion already established within $Y_{reason}$. Consequently, the log-probabilities of $Y_{answer}$ tokens can exhibit artificially high confidence, independent of the actual quality of reasoning. Including these final tokens in sequence-level evaluations hence acts as a noise factor, masking the model's true cognitive trajectory.

To extract a purer signal of the underlying thinking dynamics, we propose a reasoning-phase-isolated variant to full-sequence consilience. For any generated completion $y$, we parse the sequence to identify the semantic transition boundary and only take the reasoning portion of the generation. This isolation carries negligible cost and poses no practical obstacle: all mainstream thinking models expose an easily identifiable delimiter—e.g., \texttt{</think>} for DeepSeek-R1, Qwen3, Mistral, and Llama, and \texttt{<|channel|>final} for GPT-OSS—so the split is a single string operation. Even absent a dedicated delimiter, the \texttt{-full} variant remains a valid instantiation of the principle (Table~\ref{tab:main_results}), and any non-thinking model can be prompted to emit a "Final Answer" marker to recover a unified boundary. All subsequent boundary window calculations ($W$) and confidence aggregations are then strictly constrained to the isolated reasoning trace $Y_{reason}$, ensuring that the consilience score reflects only the genuine cognitive search and resolution process. For the non-thinking model, full sequence or segments before the "Final Answer" part are both applicable, as they enclose the actual reasoning process.

%% file: tex/exp.tex
In this section, we strive to answer the following research questions. \textbf{Q1}: How does consilience compare to other confidence-based TTS methods? \textbf{Q2}: How does consilience scale w.r.t. the number of completions? \textbf{Q3}: Does the sequence confidence follow the consilience principle? \textbf{Q4}: Is consilience robust against hyperparameters? What are the recommended hyperparameters?

\subsection{Experimental Setup}
\textbf{Model.} We utilize a diverse suite of open-weight large language models: DeepSeek-R1-Qwen-7B-Distill (DS-R1)~\citep{Guo_2025}, Qwen3-80B-Next-Instruct-FP8 (Qwen)~\citep{yang2025qwen3technicalreport}, Qwen3-80B-Coder-Next-Instruct-FP8~\citep{cao2026qwen3codernexttechnicalreport} (Qwen3-Coder-Next), and GPT-OSS-120B/20B (OSS-120B/20B)~\citep{openai2025gptoss120bgptoss20bmodel}. We use high-thinking mode for GPT-OSS-20B, and medium-thinking mode for the 120B model. This selection is intentionally designed to cover a broad spectrum of parameter sizes. Furthermore, it encompasses both explicit reasoning architectures (Deepseek and GPT-OSS) and standard instruction-tuned models, ensuring our findings generalize across different generative paradigms.

\textbf{Datasets.} We assess test-time scaling performance across four challenging, reasoning-heavy benchmarks: LiveCodeBench-v6 (LCB)~\citep{jain2024livecodebenchholisticcontaminationfree} for code generation, SWE-bench~\citep{jimenez2024swebenchlanguagemodelsresolve} for agentic application, HMMT25~\citep{balunovic2025matharena} for advanced mathematical reasoning, and GPQA-Diamond~\citep{rein2023gpqagraduatelevelgoogleproofqa} for graduate-level question answering. This selection provides comprehensive coverage across diverse cognitive domains. Crucially, LCB-v6 and SWE-bench require the generation of expressive, free-form programmatic solutions. In these environments, methods relying on internal signals like consilience are readily applicable, whereas methods relying on majority voting have limited application.

\textbf{Experimental Setup.} To assess the model's ability to scale in the general setting, we adopt the Best-of-n selection paradigm, where we sample $n$ completions, and use a test-time scaling method to rank and select the Top-1 completion as the final answer. For one-turn datasets (LCB, GPQA, HMMT), we employ a shared-pool evaluation strategy. For every query, we first generate a common pool of 256 independent completions. During evaluation, we randomly draw a subset of size $n$ from this common pool and apply each top-1 selection method. We repeat the experiment 10 times. For the agentic case (SWE-bench), we apply each method to individual steps, and we only apply test-time scaling to edit steps to save computation costs. Detailed agentic setup is in Appendix~\ref{app:agentic}. We also note that consilience is applicable to a voting-based scheme, and present voting-based results in Appendix~\ref{app:majority_voting}, where we show the robustness of consilience as a correctness signal.

\textbf{Baselines and Consilience Implementation.} Our primary baselines are Self-Certainty~\citep{kang2025scalablebestofnselectionlarge}, which calculates the mean confidence across the entire generated sequence and selects the completion that maximizes the confidence, and Deep-Conf~\cite{fu2025deepthinkconfidence}, which proposes to use only the last-2048 tokens' confidence (DeepConf-2K). We additionally add in Deep-conf-20\%, which uses the mean confidence of the last 20\% of tokens. To align with the baselines, we also use two variants, absolute tokens and dynamic percentages, to determine the initial and final window sizes ($W$) for consilience computation. We always skip the first 5\% of tokens as mentioned in Section~\ref{sec:cons_score} to focus on confidence computation. To compute token-level confidence, we use $K=5$ top log-probabilities. We choose a small $K$ for two reasons: (i) commercial inference APIs typically expose only the top-20 log-probabilities, so a small-$K$ estimator makes consilience deployable on such models; and (ii) the top candidates already carry the bulk of the probability mass—on GPT-OSS-20B/LCB the top-5 tokens capture $93.9\%$ of mass per step—so $K=5$ suffices while avoiding the large log-prob payload of full-vocabulary scoring.

\begin{table}[t]
  \centering
  \caption{Accuracy (\%) across models and datasets. All TTS methods have a sample size of $n=64$. Bold = best per column.}
  \input{tab/main_tab}
  \label{tab:main_results}
\end{table}
\begin{table}[t]
  \centering
  \caption{SWE-bench Verified issue resolve rate (\%) across models and agents. ($n=32$) Bold = best per column.}
  \input{tab/agent_result}
  \label{tab:agent_result}
\end{table}
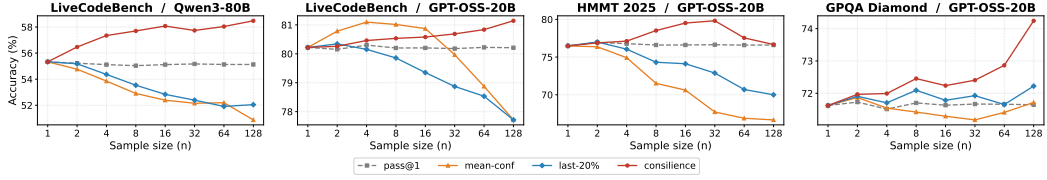
\begin{figure}[t]
  \centering
  \input{figures/fig_scaling}
  \caption{Scaling pattern on different models and datasets ($x$-axis: number of samples $n$). Consilience can scale in a robust fashion, while other metrics sometimes diminish.}
  \label{fig:scaling}
\end{figure}
\begin{figure}[t]
  \centering
  \input{figures/fig_conf_diff}
  \caption{Difference between mean confidence trajectory of correct and incorrect completions on hard problems (<50\% Pass@1). Completions show a consistent converging pattern, where the initial confidence of correct completions is lower.}
  \label{fig:conf-diff} 
\end{figure}
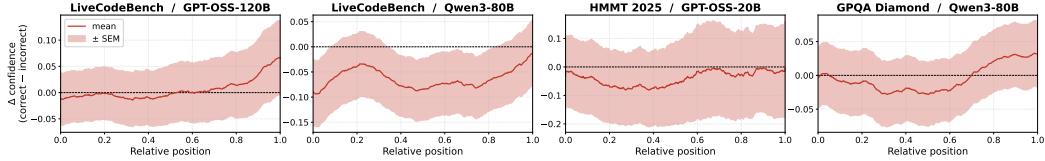

\subsection{Consilience Performance}
To answer \textbf{Q1}, Table~\ref{tab:main_results} presents the top-1 selection accuracy across the LiveCodeBench, HMMT, and GPQA datasets. We calculate the consilience score from Equation~\ref{eq:cons} using a fixed penalty multiplier of $\alpha = 3$. The boundary window itself is evaluated either dynamically as 20\% of the sequence length (Cns) or as an absolute threshold of 2048 tokens (Cns-2K), aligning with the baselines. The -think variants apply this calculation exclusively to the isolated reasoning trace, whereas -full evaluates the entire generated output as an ablation study, thinking isolation details can be found in Appendix~\ref{app:model}. For Qwen on GPQA, because the $Y_{answer}$ is just a boxed character, we only report the full version.

\textbf{The Failure of Simple Maximization.} The most surprising results empirically validate our analysis regarding the "long-tail trap" of maximization-only confidence metrics. Across multiple models and datasets, applying these metrics without voting frequently results in catastrophic performance degradation compared to random selection, Pass@1. For instance, on the HMMT benchmark, applying Self-Certainty to OSS-20B drops performance from 76.5\% down to 68.0\%. Similarly, on LCB for OSS-120B, Pass@1 achieves 65.7\%, while DeepConf-20\% degrades to 61.9\%. Without the guardrails of external verifiers, simply maximizing sequence-level confidence makes selecting confidently hallucinated, overfitted trajectories inevitable.

\textbf{Consistent Outperformance of Consilience.} In stark contrast, the consilience framework successfully neutralizes this degradation, consistently outperforming both Pass@1 and all confidence baselines across all datasets. Notably, consilience excels on the open-ended LCB code generation benchmark—a critical domain where traditional exact-match majority voting is inapplicable. On LCB, our Cns-think elevates OSS-120B to 69.7\%, Qwen from 55.3\% (Pass@1) to 60.4\%, showing necessity of early low confidence. We present in Appendix~\ref{app:qualitative} a case study comparing consilience and over-confidence examples, providing intuitve understanding of the impact of low initial confidence.

\textbf{Ablation: The Necessity of Reasoning Isolation.} The comparison between the \texttt{-full} and \texttt{-think} variants illustrates how to maximally concentrate the consilience signal. Both variants rely on the same underlying logic, evaluating the temporal asymmetry of confidence, and both are effective implementations of our framework. For instance, the full-sequence variant is remarkably strong for Qwen on LiveCodeBench. Qualitative investigation reveals that Qwen frequently continues its active problem-solving by writing reasoning steps as comments directly within the final code blocks, so the \texttt{-full} trajectory still captures rich information on model confidence. The core takeaway is that isolating the reasoning phase (\texttt{-think}) reliably elevates performance across the broader spectrum of models: by stripping away the often deterministic summarization of the final answer, isolation distills the true reasoning pattern from the trajectory. We therefore recommend \texttt{-think} as the general best practice for test-time selection.

\subsection{Agentic Application}\label{sec:agentic_exp}
To evaluate consilience in complex, multi-turn environments, we integrate it into the mini-SWE-agent framework on SWE-Bench-Verified and target TTS at critical action steps: when the model attempts a large edit step (identified via bash command matching), consilience selects the best action from $n=32$ completions sampled in parallel. Details are in Appendix~\ref{app:agentic}.

This targeted integration yields a strong improvement over the base model (Table~\ref{tab:agent_result}), showing that our temporal confidence signal reliably identifies high-quality reasoning and tool use in open-ended software engineering loops where programmatic verification is unavailable. It is also efficient: applying consilience with $n=32$ at every step would incur roughly $32\times$ the computation, whereas with our harness Qwen3 takes 18 hours versus 8 hours for the plain mini-SWE-agent on our hardware. Smarter harnesses that trigger consilience only at the highest-value steps could reduce this cost further.

\begin{table}[t]
  \centering
  \begin{minipage}[t]{0.42\textwidth}
  \centering
  \caption{Difficulty-stratified selection accuracy (\%) on LiveCodeBench, split into equally-sized tiers by Pass@1 rate. Consilience is neutral on easy problems and improves on medium/hard tiers.}
  \resizebox{\linewidth}{!}{\input{tab/difficulty}}
  \label{tab:difficulty}
  \end{minipage}\hfill
  \begin{minipage}[t]{0.54\textwidth}
  \centering
  \caption{Hyperparameter transfer. \emph{Suggested} uses the frozen $\alpha=3, k=20\%$; \emph{Selected} is the per-set grid-search optimum under 5-fold cross-validation. The last row transfers parameters chosen on LCB to HMMT.}
  \resizebox{\linewidth}{!}{\input{tab/transfer}}
  \label{tab:transfer}
  \end{minipage}
\end{table}

\subsection{Investigation on Consilience}\label{sec:investigation}
\textbf{Scaling Robustness.} To answer \textbf{Q2}, Figure~\ref{fig:scaling} illustrates the accuracy scaling behavior as the sample size ($n$) increases up to 128. Consilience generally exhibits superior and more consistent scaling properties compared to mean-conf (Self-Certainty) or last-20\% (DeepConf-20\%). On GPQA (OSS-20B), while both consilience and the last-20\% scale positively, consilience scales at a visibly steeper rate. The advantage is most pronounced on free-form generation tasks (LCB). Using Qwen, only consilience scales positively. With OSS-20B, while the mean-conf baseline scales well initially, it quickly falls into the long-tail trap, suffering severe degradation as $n$ increases, a similar trend observed in~\citet{chen2024llmcallsneedscaling}. In contrast, consilience scales steadily through $n=128$. While consilience occasionally exhibits minor degradation on HMMT for OSS-20B, it consistently preserves performance above the Pass@1 floor.

\textbf{Temporal Confidence Trajectories.} To answer \textbf{Q3} and validate the principle of consilience, Figure~\ref{fig:conf-diff} normalizes the confidence within each problem, resamples confidence trajectories to the same length, plots the mean difference trajectory (correct - incorrect) on hard problems with Pass@1 rate < 50\%. Across all evaluated models and datasets, the difference is negative or lowest at the beginning of the sequence and steadily rises toward the end. This empirically confirms that correct reasoning trajectories inherently possess lower initial confidence and higher relative final confidence. In some environments, the correct trajectories maintain a lower absolute confidence, further proving that an unpenalized high-confidence state actively correlates with over-confidence rather than reasoning.

\textbf{Statistical Significance.} This trajectory difference is statistically reliable rather than an artifact of the mean. On hard problems (Pass@1 $<20\%$), consilience separates correct from incorrect completions with AUROC 0.61--0.62 (non-overlapping 95\% CIs), whereas mean confidence is at chance (0.47--0.50). A paired test of the within-problem rise $\Delta = C_{final} - C_{initial}$ is significant on both LiveCodeBench panels ($p=0.011, 0.018$; $d_z\approx0.35$) and positive on GPQA (Wilcoxon $p=0.042$). Full tables and test details are in Appendix~\ref{app:stat}.

\textbf{Difficulty Stratification.} The trajectory analysis predicts that consilience should help precisely where mean-confidence maximization fails. We verify this by splitting each dataset into three equally-sized difficulty tiers by Pass@1 rate and reporting mean-confidence versus consilience selection accuracy in Table~\ref{tab:difficulty}. On easy problems, Pass@1 is already near-perfect and consilience is neutral—matching the confidence baseline (99.9 vs.\ 99.9 for Qwen; 99.1 vs.\ 100.0 for GPT-OSS-120B) because all completions in the pool exhibit similarly high initial confidence, leaving no signal to penalize. The gains concentrate on medium and hard problems (up to $+13.7$ on medium), precisely where premature convergence is the dominant failure mode. This confirms that the initial-confidence penalty is harmless where confidence maximization already succeeds, and beneficial where it does not, directly addressing when high prefix confidence should be rewarded rather than penalized.

\begin{figure}[t]
  \centering
  \input{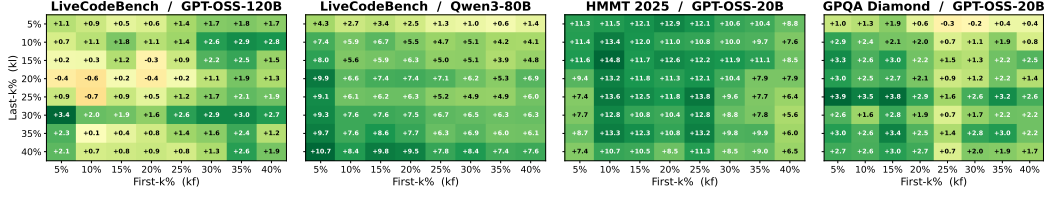}
  \caption{Difference between Consilience Accuracy (last window=$k_l\%$ and first window = $k_f\%$) and last-$k_l\%$ accuracy. Penalizing initial confidence is consistently beneficial.} 
  \label{fig:heatmap}
\end{figure}

\begin{figure}[t]
  \centering
  \input{figures/fig_alpha}
  \caption{Consilience performance with different $\alpha$. Stars mark the maximum performance improvement over $\alpha=0$, and $\alpha>0$ systematically brings performance boosts.}
  \label{fig:alpha-sweep}
\end{figure}
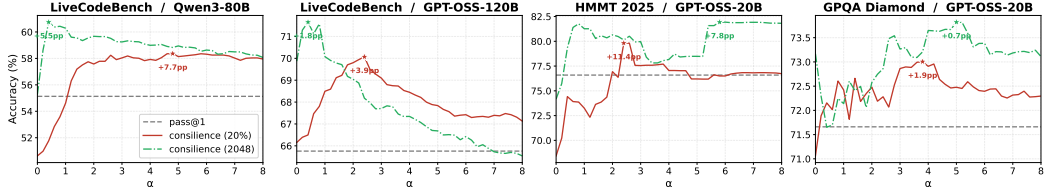

\subsection{Robustness of Consilience} \label{sec:robust}
We answer \textbf{Q4} by analyzing sensitivity to the two primary hyperparameters: the boundary window sizes and the penalty multiplier $\alpha$.

\textbf{Window Size Robustness.} While we use the same window size in Section~\ref{sec:cons_score}, consilience can naturally be extended to different initial and last window sizes. We evaluated the performance delta between using our consilience score and using the corresponding last-k\% across varying initial and final window percentages in Figure~\ref{fig:heatmap}. The resulting heatmaps reveal that incorporating the initial confidence penalty consistently yields a positive performance boost across a massive combinatorial range of percentages. A generic robust range across model and data is 20\% to 30\%.

\textbf{The Penalty Multiplier ($\alpha$).} We sweep $\alpha$ from 0 to 8 in Figure~\ref{fig:alpha-sweep}. A proper $\alpha > 0$ yields a positive gain that consistently outperforms the pass@1 baseline, and we advise $\alpha \in [2, 5]$. A weight above 1 is needed because autoregressive confidence naturally starts low and ends high, so the initial-exploration penalty must be scaled up to balance the final convergence score.
Performance degrades when $\alpha$ becomes excessively large, which illustrates the dual nature of our theory: early exploration is necessary, but final convergence is equally critical. Over-penalizing initial confidence ignores convergence and selects from both Category 3 (consilience) and Category 4 (total confusion) in Table~\ref{tab:cognitive_states}.

\textbf{Selection Protocol and Transfer.} To make the selection protocol explicit and rule out per-benchmark tuning, we fixed $\alpha=3$, window $k=20\%$, and the $5\%$ skipped prefix on a single development setting (Qwen on LiveCodeBench) \emph{before} evaluating any other benchmark; these frozen values are the reported \texttt{Cns} configuration in Table~\ref{tab:main_results}. To confirm these are not over-tuned, we additionally run a grid search ($\alpha\in\{1,\dots,5\}$, $k\in\{5\%,\dots,30\%\}$) under 5-fold cross-validation—selecting hyperparameters on the training fold and evaluating on the held-out fold—and a cross-dataset transfer where parameters chosen on GPT-OSS-20B/LCB are applied unchanged to HMMT (Table~\ref{tab:transfer}). The frozen configuration captures 72--78\% of the per-set optimal gain, and transfers across datasets without re-tuning (+2.2 over Pass@1 on HMMT), suggesting a practical recipe: curate a small seed set per model to fix hyperparameters, then apply them to unseen data.

%% file: tab/main_tab.tex
\resizebox{0.99\textwidth}{!}{
\begin{tabular}{lcccccccc}
    \toprule
    & \multicolumn{4}{c}{\textbf{LCB}} & \multicolumn{2}{c}{\textbf{HMMT}} & \multicolumn{2}{c}{\textbf{GPQA}} \\
    \cmidrule(lr){2-5} \cmidrule(lr){6-7} \cmidrule(lr){8-9}
    \textbf{Method} & GPT-OSS-120B & GPT-OSS-20B & DS-R1 & Qwen & GPT-OSS-20B & Qwen & GPT-OSS-20B & Qwen \\
    \midrule
    Pass@1 & 65.7{\scriptsize$\pm$0.2} & 80.2{\scriptsize$\pm$0.1} & 47.8{\scriptsize$\pm$0.1} & 55.3{\scriptsize$\pm$0.2} & 76.5{\scriptsize$\pm$0.3} & 52.6{\scriptsize$\pm$0.5} & 71.8{\scriptsize$\pm$0.2} & 60.5{\scriptsize$\pm$0.2} \\
    Self-Certainty & 64.5{\scriptsize$\pm$1.3} & 79.0{\scriptsize$\pm$1.3} & 54.9{\scriptsize$\pm$1.5} & 53.2{\scriptsize$\pm$1.7} & 68.0{\scriptsize$\pm$3.1} & 45.3{\scriptsize$\pm$3.1} & 71.7{\scriptsize$\pm$1.4} & 57.8{\scriptsize$\pm$1.8} \\
    DeepConf-20\% & 61.9{\scriptsize$\pm$1.3} & 79.1{\scriptsize$\pm$0.9} & 53.7{\scriptsize$\pm$1.6} & 52.0{\scriptsize$\pm$1.6} & 69.3{\scriptsize$\pm$3.3} & 51.3{\scriptsize$\pm$4.3} & 71.0{\scriptsize$\pm$1.2} & 60.4{\scriptsize$\pm$1.3} \\
    DeepConf-2K & 62.5{\scriptsize$\pm$0.8} & 76.7{\scriptsize$\pm$1.2} & 53.7{\scriptsize$\pm$1.4} & 55.4{\scriptsize$\pm$1.8} & 73.3{\scriptsize$\pm$4.9} & 47.3{\scriptsize$\pm$3.6} & 71.3{\scriptsize$\pm$1.3} & 57.6{\scriptsize$\pm$2.4} \\ \midrule
    Cns-full & 62.2{\scriptsize$\pm$1.0} & 80.2{\scriptsize$\pm$0.4} & 52.0{\scriptsize$\pm$1.9} & 59.5{\scriptsize$\pm$2.0} & 78.3{\scriptsize$\pm$2.7} & 49.7{\scriptsize$\pm$3.5} & 72.5{\scriptsize$\pm$1.1} & \textbf{65.2{\scriptsize$\pm$1.6}} \\
    Cns-2K-full & 64.6{\scriptsize$\pm$1.7} & 79.4{\scriptsize$\pm$0.8} & 56.0{\scriptsize$\pm$0.3} & \textbf{60.9{\scriptsize$\pm$1.6}} & 77.0{\scriptsize$\pm$4.6} & 51.0{\scriptsize$\pm$3.3} & 73.8{\scriptsize$\pm$1.4} & 58.1{\scriptsize$\pm$1.5} \\ \midrule
    Cns-think & \textbf{69.7{\scriptsize$\pm$0.9}} & \textbf{81.1{\scriptsize$\pm$0.7}} & 54.0{\scriptsize$\pm$0.5} & 58.7{\scriptsize$\pm$0.9} & 78.0{\scriptsize$\pm$2.7} & 52.0{\scriptsize$\pm$2.7} & 72.0{\scriptsize$\pm$1.0} & - \\
    Cns-2K-think & 67.8{\scriptsize$\pm$1.5} & 80.7{\scriptsize$\pm$0.6} & \textbf{57.1{\scriptsize$\pm$0.9}} & 60.4{\scriptsize$\pm$1.9} & \textbf{80.7{\scriptsize$\pm$2.5}} & \textbf{53.7{\scriptsize$\pm$2.3}} & \textbf{73.8{\scriptsize$\pm$1.5}} & - \\
    \bottomrule
  \end{tabular}}

%% file: tab/agent_result.tex
\begin{tabular}{llll}
\toprule
 Tool         & Model              & \% Resolved   \\ \hline
 mini-SWE-agent      & GPT-OSS-120B &    23.0{\scriptsize$\pm$0.8}   \\
 mini-SWE-agent + Consilience     & GPT-OSS-120B &  \textbf{26.9{\scriptsize$\pm$1.1}}  \\ \midrule
 mini-SWE-agent      & Qwen3-Coder-Next &  65.3{\scriptsize$\pm$0.8}      \\
 mini-SWE-agent + Consilience     & Qwen3-Coder-Next & \textbf{67.3{\scriptsize$\pm$1.2}}     
     \\
\bottomrule
\end{tabular}

%% file: figures/fig_scaling.tex
\begin{tikzpicture}
\begin{groupplot}[
  group style={group size=4 by 1, horizontal sep=0.78cm},
  cs panel, cs legend,
  width=2.63cm, height=1.15cm,
  xmode=log, log basis x=2,
  xmin=0.8, xmax=160,
  xtick={1,2,4,8,16,32,64,128},
  xticklabels={1,,4,,16,,64,},
]
\nextgroupplot[
  title={LCB, Qwen3-80B},
  ymin=50.3, ymax=59.2, ytick={52,54,56,58},
  ylabel={Accuracy (\%)},
  legend style={at={(axis description cs:0,1.24)}, anchor=south west},
]
\addplot[cs line, csBase, line width=0.7pt] table[x=n, y=pass1] {figures/data/scaling_1.dat};
\addlegendentry{pass@1}
\addplot[cs line, csOver, cs dot=csOver] table[x=n, y=mean] {figures/data/scaling_1.dat};
\addlegendentry{Mean confidence (Self-Certainty)}
\addplot[cs line, csLast, cs dot=csLast] table[x=n, y=last20] {figures/data/scaling_1.dat};
\addlegendentry{Last-20\% confidence (DeepConf)}
\addplot[cs line, csCons, cs dot=csCons] table[x=n, y=cons] {figures/data/scaling_1.dat};
\addlegendentry{Consilience}
\nextgroupplot[title={LCB, GPT-OSS-20B}, ymin=77.4, ymax=81.5, ytick={78,79,80,81}]
\addplot[cs line, csBase, line width=0.7pt] table[x=n, y=pass1] {figures/data/scaling_2.dat};
\addplot[cs line, csOver, cs dot=csOver] table[x=n, y=mean] {figures/data/scaling_2.dat};
\addplot[cs line, csLast, cs dot=csLast] table[x=n, y=last20] {figures/data/scaling_2.dat};
\addplot[cs line, csCons, cs dot=csCons] table[x=n, y=cons] {figures/data/scaling_2.dat};
\nextgroupplot[title={HMMT, GPT-OSS-20B}, ymin=65.5, ymax=81, ytick={70,75,80}]
\addplot[cs line, csBase, line width=0.7pt] table[x=n, y=pass1] {figures/data/scaling_3.dat};
\addplot[cs line, csOver, cs dot=csOver] table[x=n, y=mean] {figures/data/scaling_3.dat};
\addplot[cs line, csLast, cs dot=csLast] table[x=n, y=last20] {figures/data/scaling_3.dat};
\addplot[cs line, csCons, cs dot=csCons] table[x=n, y=cons] {figures/data/scaling_3.dat};
\nextgroupplot[title={GPQA, GPT-OSS-20B}, ymin=70.9, ymax=74.6, ytick={71,72,73,74}]
\addplot[cs line, csBase, line width=0.7pt] table[x=n, y=pass1] {figures/data/scaling_4.dat};
\addplot[cs line, csOver, cs dot=csOver] table[x=n, y=mean] {figures/data/scaling_4.dat};
\addplot[cs line, csLast, cs dot=csLast] table[x=n, y=last20] {figures/data/scaling_4.dat};
\addplot[cs line, csCons, cs dot=csCons] table[x=n, y=cons] {figures/data/scaling_4.dat};
\end{groupplot}
\end{tikzpicture}

%% file: figures/fig_conf_diff.tex
\begin{tikzpicture}
\begin{groupplot}[
  group style={group size=4 by 1, horizontal sep=0.85cm},
  cs panel,
  width=2.43cm, height=1.3cm,
  xmin=0, xmax=1, xtick={0,0.5,1}, xticklabels={start,,end},
]
\nextgroupplot[title={LCB, GPT-OSS-120B}, ymin=-0.07, ymax=0.14, ytick={-0.05,0,0.05,0.1},
  yticklabels={$-$0.05,0,0.05,0.10}, ylabel={$\Delta$ confidence}]
\addplot[name path=hi, draw=none, forget plot] table[x=x, y=hi] {figures/data/confdiff_1.dat};
\addplot[name path=lo, draw=none, forget plot] table[x=x, y=lo] {figures/data/confdiff_1.dat};
\addplot[fill=csCons, fill opacity=0.15, draw=none, forget plot] fill between[of=hi and lo];
\addplot[csInkTwo, line width=0.5pt, forget plot] coordinates {(0,0) (1,0)};
\addplot[cs line, csCons] table[x=x, y=mean] {figures/data/confdiff_1.dat};

\nextgroupplot[title={LCB, Qwen3-80B}, ymin=-0.165, ymax=0.06, ytick={-0.15,-0.1,-0.05,0,0.05},
  yticklabels={$-$0.15,$-$0.10,$-$0.05,0,0.05}]
\addplot[name path=hi, draw=none, forget plot] table[x=x, y=hi] {figures/data/confdiff_2.dat};
\addplot[name path=lo, draw=none, forget plot] table[x=x, y=lo] {figures/data/confdiff_2.dat};
\addplot[fill=csCons, fill opacity=0.15, draw=none, forget plot] fill between[of=hi and lo];
\addplot[csInkTwo, line width=0.5pt, forget plot] coordinates {(0,0) (1,0)};
\addplot[cs line, csCons] table[x=x, y=mean] {figures/data/confdiff_2.dat};

\nextgroupplot[title={HMMT, GPT-OSS-20B}, ymin=-0.22, ymax=0.17, ytick={-0.2,-0.1,0,0.1},
  yticklabels={$-$0.2,$-$0.1,0,0.1}]
\addplot[name path=hi, draw=none, forget plot] table[x=x, y=hi] {figures/data/confdiff_3.dat};
\addplot[name path=lo, draw=none, forget plot] table[x=x, y=lo] {figures/data/confdiff_3.dat};
\addplot[fill=csCons, fill opacity=0.15, draw=none, forget plot] fill between[of=hi and lo];
\addplot[csInkTwo, line width=0.5pt, forget plot] coordinates {(0,0) (1,0)};
\addplot[cs line, csCons] table[x=x, y=mean] {figures/data/confdiff_3.dat};

\nextgroupplot[title={GPQA, Qwen3-80B}, ymin=-0.08, ymax=0.085, ytick={-0.05,0,0.05},
  yticklabels={$-$0.05,0,0.05}]
\addplot[name path=hi, draw=none, forget plot] table[x=x, y=hi] {figures/data/confdiff_4.dat};
\addplot[name path=lo, draw=none, forget plot] table[x=x, y=lo] {figures/data/confdiff_4.dat};
\addplot[fill=csCons, fill opacity=0.15, draw=none, forget plot] fill between[of=hi and lo];
\addplot[csInkTwo, line width=0.5pt, forget plot] coordinates {(0,0) (1,0)};
\addplot[cs line, csCons] table[x=x, y=mean] {figures/data/confdiff_4.dat};
\end{groupplot}
\end{tikzpicture}

%% file: tab/difficulty.tex
\begin{tabular}{llccc}
    \toprule
    \textbf{Model} & \textbf{Method} & \textbf{Easy} & \textbf{Medium} & \textbf{Hard} \\
    \midrule
    \multirow{2}{*}{Qwen} & Mean-conf & 99.9 & 58.1 & 3.7 \\
     & Consilience & 99.9 & \textbf{70.3} & \textbf{8.1} \\
    \midrule
    \multirow{2}{*}{GPT-OSS-120B} & Mean-conf & 100.0 & 74.9 & 13.4 \\
     & Consilience & 99.1 & \textbf{88.6} & \textbf{17.2} \\
    \bottomrule
\end{tabular}

%% file: tab/transfer.tex
\begin{tabular}{lccc}
    \toprule
    \textbf{Setting} & \textbf{Pass@1} & \textbf{Suggested} & \textbf{Selected} \\
    \midrule
    GPT-OSS-120B / LCB & 65.3 & 70.2 & 72.8 \\
    Qwen / LCB & 54.9 & 58.1 & 61.3 \\
    GPT-OSS-20B / LCB & 80.2 & 81.3 & 83.7 \\
    \midrule
    GPT-OSS-20B / LCB $\rightarrow$ HMMT & 75.9 & 78.1 & 80.2 \\
    \bottomrule
\end{tabular}

%% file: figures/fig_alpha.tex
\pgfplotsset{
  cs star/.style={only marks, mark=star, mark size=2.1pt, mark options={draw=#1, fill=#1, line width=0.7pt}, forget plot},
}
\tikzset{cs gain/.style={cs note, fill=white, fill opacity=0.85, text opacity=1, inner sep=0.6pt}}
\begin{tikzpicture}
\begin{groupplot}[
  group style={group size=4 by 1, horizontal sep=0.8cm},
  cs panel, cs legend,
  width=2.56cm, height=1.42cm,
  xmin=0, xmax=8, xtick={0,2,4,6,8},
]
\nextgroupplot[
  title={LCB, Qwen3-80B}, ymin=50, ymax=62, ytick={52,56,60},
  ylabel={Accuracy (\%)},
  legend style={at={(axis description cs:2.469,1.22)}, anchor=south},
]
\addplot[cs line, csCons] table[x=alpha, y=acc] {figures/data/alpha_pct_1.dat};
\addlegendentry{Consilience, 20\% window}
\addplot[cs line, csConsDark, densely dashed] table[x=alpha, y=acc] {figures/data/alpha_tok_1.dat};
\addlegendentry{Consilience, 2048-token window}
\addplot[cs line, csBase, line width=0.7pt] coordinates {(0,55.13) (8,55.13)};
\addlegendentry{pass@1}
\addplot[cs star=csCons] coordinates {(4.8,58.36)};
\addplot[cs star=csConsDark] coordinates {(0.4,60.74)};
\node[cs gain, anchor=north] at (axis cs:4.8,57.9) {+7.7};
\node[cs gain, anchor=west] at (axis cs:0.7,61.1) {+5.5};
\nextgroupplot[title={LCB, GPT-OSS-120B}, ymin=65.2, ymax=72.4, ytick={66,68,70,72}]
\addplot[cs line, csCons] table[x=alpha, y=acc] {figures/data/alpha_pct_2.dat};
\addplot[cs line, csConsDark, densely dashed] table[x=alpha, y=acc] {figures/data/alpha_tok_2.dat};
\addplot[cs line, csBase, line width=0.7pt] coordinates {(0,65.76) (8,65.76)};
\addplot[cs star=csCons] coordinates {(2.4,70.07)};
\addplot[cs star=csConsDark] coordinates {(0.4,71.65)};
\node[cs gain, anchor=south west] at (axis cs:2.6,70.1) {+3.9};
\node[cs gain, anchor=west] at (axis cs:0.7,71.75) {+1.8};
\nextgroupplot[title={HMMT, GPT-OSS-20B}, ymin=67.5, ymax=83.5, ytick={70,75,80}]
\addplot[cs line, csCons] table[x=alpha, y=acc] {figures/data/alpha_pct_3.dat};
\addplot[cs line, csConsDark, densely dashed] table[x=alpha, y=acc] {figures/data/alpha_tok_3.dat};
\addplot[cs line, csBase, line width=0.7pt] coordinates {(0,76.59) (8,76.59)};
\addplot[cs star=csCons] coordinates {(2.4,79.82)};
\addplot[cs star=csConsDark] coordinates {(5.8,81.93)};
\node[cs gain, anchor=east] at (axis cs:2.2,78.2) {+11.4};
\node[cs gain, anchor=east] at (axis cs:5.55,82.3) {+7.8};
\nextgroupplot[title={GPQA, GPT-OSS-20B}, ymin=70.9, ymax=74.5, ytick={71,72,73,74}]
\addplot[cs line, csCons] table[x=alpha, y=acc] {figures/data/alpha_pct_4.dat};
\addplot[cs line, csConsDark, densely dashed] table[x=alpha, y=acc] {figures/data/alpha_tok_4.dat};
\addplot[cs line, csBase, line width=0.7pt] coordinates {(0,71.66) (8,71.66)};
\addplot[cs star=csCons] coordinates {(3.8,73.01)};
\addplot[cs star=csConsDark] coordinates {(5.0,73.82)};
\node[cs gain, anchor=north east] at (axis cs:3.7,72.9) {+1.9};
\node[cs gain, anchor=south] at (axis cs:5.0,73.92) {+0.7};
\end{groupplot}
\end{tikzpicture}

%% file: tex/app_dataset.tex
\label{app:datasets}
To rigorously evaluate the efficacy of the consilience metric, we selected a diverse suite of benchmarks encompassing advanced mathematics, graduate-level science, and open-ended software engineering. These datasets were specifically chosen for their high difficulty and, critically, their requirement for complex, multi-step reasoning where standard heuristics often fail. Our script and agentic implementation can be found here: \href{https://github.com/LechengKong/consilience}{https://github.com/LechengKong/consilience}.

\begin{itemize}
\item \textbf{HMMT25-Feb}~\citep{balunovic2025matharena}: The Harvard-MIT Mathematics Tournament (February 2025) comprises exceptionally difficult competition-level mathematics problems. It demands rigorous, multi-step logical deductions and creative problem-solving, making it a robust testbed for advanced mathematical reasoning. We use the provided script in MathArena benchmark to compare and group generated Latex answers.
\item \textbf{AIME25}~\citep{balunovic2025matharena}: The 2025 American Invitational Mathematics Examination consists of highly challenging, integer-answer mathematics problems. It rigorously evaluates a model's capacity for extended algebraic, geometric, and combinatorial reasoning under strict constraints. We use the provided script in MathArena benchmark to compare and group generated Latex answers.
\item \textbf{GPQA-Diamond}~\citep{rein2023gpqagraduatelevelgoogleproofqa}: GPQA is a highly challenging dataset of graduate-level questions in physics, biology, and chemistry. The "Diamond" subset is specifically curated to be difficult even for highly skilled human experts, providing a rigorous evaluation of deep domain knowledge and complex scientific reasoning.
\item \textbf{LiveCodeBench-v6}~\citep{jain2024livecodebenchholisticcontaminationfree}: A dynamic code generation benchmark utilizing recent contest problems (e.g., from LeetCode, AtCoder) to mitigate training data contamination. It requires models to generate expressive, free-form programmatic solutions, serving as our primary environment to evaluate selection metrics where exact-match majority voting is inapplicable. For this LiveCodeBench, we use only problem from v6 release, excluding problems from v1-v5 release to control the effect of data contamination and focus on the harder problems. We use the evalutaion script supplied with the benchmark to evaluate the results.
\item \textbf{SWE-bench-Verified}~\citep{jimenez2024swebenchlanguagemodelsresolve}: A framework for evaluating large language models on real-world software engineering tasks. It requires models to act as autonomous agents, navigating complex code repositories and generating multi-step, open-ended edits to resolve actual GitHub issues. We use the minimal bash-only agent, mini-SWE-agent~\citep{yang2024sweagent}, provided along with the benchmark to focus on evaluating the power of consilience.
\item \textbf{Natural Questions}~\citep{kwiatkowski2019natural}: An open-domain common-sense question answering benchmark of real Google search queries with answers grounded in Wikipedia. We use it (a 2K-question subsample) to evaluate consilience on a data distribution outside the mathematics and code domains; see Appendix~\ref{app:nq}.
\end{itemize}

%% file: tex/app_model.tex
\label{app:model}
All model inference is conducted using the vLLM framework hosted on a node with 4$\times$ NVIDIA L40S (48GB) GPUs. Unless otherwise specified, we utilize the default generation configurations provided by the models' respective HuggingFace repositories. The specific sampling parameters and reasoning effort configurations are detailed in Table~\ref{tab:model_configs}.

\begin{table}[h]
\centering
\caption{Sampling configurations and thinking parameters for the evaluated models. Default Hugging Face generation parameters were used where applicable.}
\begin{tabular}{lcccccc}
\toprule
\textbf{Model} & \textbf{Temperature} & \textbf{Top-$p$} & \textbf{Top-$k$} & \textbf{Max Len.} & \textbf{Reasoning Effort} \\
\midrule
GPT-OSS-120B & 1.0 & 1.0 & 40 & 130k & Medium \\
GPT-OSS-20B & 1.0 & 1.0 & 40 & 130k & High \\
Qwen3-Next-80B & 0.7 & 0.8 & 20 & 130k & N/A \\
DeepSeek-R1-Qwen-7B & 0.6 & 0.95 & 40 & 130k & N/A \\
\bottomrule
\end{tabular}
\vspace{0.1cm}

\label{tab:model_configs}
\end{table}

Due to computational resource constraints, we restrict the "high" reasoning effort mode to the GPT-OSS-20B model, while utilizing the "medium" setting for the larger 120B variant. This variation in configuration also serves to demonstrate the robust applicability of the consilience metric across distinctly different thinking patterns and generation lengths. We observe that under the "high" thinking mode, GPT-OSS-20B occasionally fails to complete its answer within the 130k maximum token budget. To ensure a strictly fair evaluation, any such truncated responses are filtered out of the generation pool prior to computing Pass@1 and evaluating all other selection baselines.

\textbf{Implementation of Reasoning Isolation.} To execute the reasoning-phase isolation, we employ model-specific structural delimiters to parse the reasoning trace ($Y_{reason}$) from the final answer ($Y_{answer}$). For the GPT-OSS models, we extract tokens preceding the \texttt{<|channel|>final} delimiter. For DeepSeek-R1, we truncate the sequence at the standard \texttt{</think>} tag. For Qwen3-Next, which lacks a dedicated end-of-thought token in its instruction format, we apply task-specific structural heuristics: on LiveCodeBench, we extract the sequence prior to the first \texttt{\textasciigrave\textasciigrave\textasciigrave python} block; on HMMT, we split the trace at the first occurrence of the \texttt{\textbackslash boxed} macro, because the model generate large trunk of justification texts and reiterates on the answer; on GPQA, and AIME, we use the full trajectory, because the answer is just a single interger or character; on Swe-bench, we split the trace by \texttt{\textasciigrave\textasciigrave\textasciigrave bash}. We use these specialized delimiters because we are following the original prompt provided by the benchmark. In reality or in production, one can always prompt non-thinking model to output a "Final Answer" token before the answer, which makes thinking isolation simple and unified.

%% file: tex/app_majority_vote.tex
\label{app:majority_voting}
\begin{table}[t]
    \centering
    \caption{Accuracy (\%) of confidence metrics integrated with Borda Count voting.}
    \input{tab/voting.tex}
    \label{tab:voting_results}
\end{table}
While the primary focus of the consilience metric is to enable verifier-free selection in open-ended domains where exact-match consensus is impossible, it can also be seamlessly integrated with traditional majority voting techniques on structured tasks. To evaluate this, we combine the confidence metrics with voting on the HMMT, GPQA, and AIME datasets, where answers can be strictly extracted and matched.

Because different confidence metrics (e.g., DeepConf vs. Consilience) operate on entirely different mathematical scales and distributions, directly aggregating their raw scores is unstable. To ensure a fair and robust comparison, we utilize \textbf{Borda Count voting}. Under this scheme, instead of aggregating raw scores, each metric ranks the $N$ sampled completions. For a pool of $N$ candidates, the top-ranked completion receives $N$ points, the second receives $N-1$ points, and so on. A hyperparameter $p$ controls the exponent of the rank weighting during aggregation. 

Specifically, for a set of $N$ sampled completions $\{y_1, y_2, \dots, y_N\}$ for a given prompt, let $R(y_i) \in \{1, \dots, N\}$ denote the rank of completion $y_i$ based on the chosen selection metric (where $1$ represents the highest metric score). Let $A(y_i)$ represent the extracted final answer from completion $y_i$. The total Borda score $B(a)$ for a unique answer candidate $a$ is computed as:

\begin{equation}B(a) = \sum_{i=1}^{N} \mathbb{I}(A(y_i) = a) \cdot (N - R(y_i) + 1)^p\end{equation}

where $\mathbb{I}$ is the indicator function. The final selected answer is simply the candidate that maximizes this score: $\argmax_a B(a)$.

The results in Table~\ref{tab:voting_results} demonstrate that under a voting scheme, almost all test-time scaling methods successfully and heavily outperform the greedy Pass@1 baseline. Because the standard Self-Consistency baseline is already exceptionally high on these structured, closed-form tasks (e.g., 96.7\% on AIME for OSS-20B), the margins between the various confidence metrics become naturally compressed.

Despite this ceiling effect, the consilience metric still manages to slightly outperform the baselines, achieving the peak scores across almost all model-dataset pairs. This indicates that even when exact-match consensus is available, the temporal trajectory evaluated by consilience provides a superior, additive signal of reasoning quality. However, we emphasize that while consilience is still competitive in these environments, its primary structural advantage remains its ability to unlock these exact types of scaling margins in free-form, un-votable domains like LiveCodeBench and SWE-bench.

%% file: tab/voting.tex
\begin{tabular}{lccccccc}
    \toprule
    & & \multicolumn{2}{c}{\textbf{HMMT}} & \multicolumn{2}{c}{\textbf{GPQA}} & \multicolumn{2}{c}{\textbf{AIME}} \\
    \cmidrule(lr){3-4} \cmidrule(lr){5-6} \cmidrule(lr){7-8}
    \textbf{Method} & \textbf{p} & 20B & Qwen & 20B & Qwen & 20B & Qwen \\
    \midrule
    Pass@1 & — & 76.6{\scriptsize$\pm$0.3} & 52.5{\scriptsize$\pm$0.2} & 71.7{\scriptsize$\pm$0.2} & 60.5{\scriptsize$\pm$0.1} & 93.0{\scriptsize$\pm$0.0} & 69.4{\scriptsize$\pm$0.6} \\
    Self-Consistency & 0 & 92.7{\scriptsize$\pm$1.3} & 57.0{\scriptsize$\pm$1.0} & 90.3{\scriptsize$\pm$0.2} & 68.0{\scriptsize$\pm$0.9} & 96.7{\scriptsize$\pm$0.0} & 76.0{\scriptsize$\pm$1.3} \\
    \midrule
    Self-Certainty & 0.5 & \textbf{93.3{\scriptsize$\pm$0.0}} & 57.3{\scriptsize$\pm$1.3} & 90.2{\scriptsize$\pm$0.3} & 67.9{\scriptsize$\pm$1.5} & 96.7{\scriptsize$\pm$0.0} & 73.7{\scriptsize$\pm$1.0} \\
     & 1 & 90.0{\scriptsize$\pm$0.0} & 57.0{\scriptsize$\pm$2.3} & 90.1{\scriptsize$\pm$0.2} & 67.9{\scriptsize$\pm$1.1} & 96.7{\scriptsize$\pm$0.0} & 73.3{\scriptsize$\pm$0.0} \\
     & 2 & 90.0{\scriptsize$\pm$0.0} & 54.7{\scriptsize$\pm$3.4} & 89.7{\scriptsize$\pm$0.2} & 68.1{\scriptsize$\pm$1.2} & 93.3{\scriptsize$\pm$0.0} & 73.7{\scriptsize$\pm$1.0} \\
     & 4 & 90.0{\scriptsize$\pm$0.0} & 53.0{\scriptsize$\pm$3.8} & 89.0{\scriptsize$\pm$0.3} & 66.6{\scriptsize$\pm$1.5} & 93.3{\scriptsize$\pm$0.0} & 75.0{\scriptsize$\pm$1.7} \\
    \midrule
    DeepConf-2K & 0.5 & 89.7{\scriptsize$\pm$1.0} & 58.0{\scriptsize$\pm$1.6} & \textbf{90.3{\scriptsize$\pm$0.2}} & 67.9{\scriptsize$\pm$1.5} & 96.7{\scriptsize$\pm$0.0} & 74.3{\scriptsize$\pm$1.5} \\
     & 1 & 90.0{\scriptsize$\pm$0.0} & 58.7{\scriptsize$\pm$1.6} & \textbf{90.3{\scriptsize$\pm$0.2}} & 67.5{\scriptsize$\pm$1.4} & 96.7{\scriptsize$\pm$0.0} & 74.3{\scriptsize$\pm$2.1} \\
     & 2 & 90.0{\scriptsize$\pm$0.0} & 59.0{\scriptsize$\pm$1.5} & 90.2{\scriptsize$\pm$0.2} & 67.6{\scriptsize$\pm$1.4} & 96.7{\scriptsize$\pm$0.0} & 77.0{\scriptsize$\pm$2.3} \\
     & 4 & 86.3{\scriptsize$\pm$1.0} & 57.0{\scriptsize$\pm$1.8} & 89.4{\scriptsize$\pm$0.4} & 66.2{\scriptsize$\pm$1.5} & 96.7{\scriptsize$\pm$0.0} & 77.0{\scriptsize$\pm$2.3} \\
    \midrule
    DeepConf-20\% & 0.5 & 90.0{\scriptsize$\pm$0.0} & 57.7{\scriptsize$\pm$1.5} & \textbf{90.3{\scriptsize$\pm$0.2}} & 68.5{\scriptsize$\pm$1.2} & 96.7{\scriptsize$\pm$0.0} & 75.3{\scriptsize$\pm$1.6} \\
     & 1 & 90.0{\scriptsize$\pm$0.0} & 58.7{\scriptsize$\pm$1.6} & 90.1{\scriptsize$\pm$0.2} & 68.3{\scriptsize$\pm$1.2} & 96.7{\scriptsize$\pm$0.0} & 76.0{\scriptsize$\pm$3.3} \\
     & 2 & 90.0{\scriptsize$\pm$0.0} & \textbf{59.3{\scriptsize$\pm$1.3}} & 89.1{\scriptsize$\pm$0.5} & 67.9{\scriptsize$\pm$1.0} & 96.7{\scriptsize$\pm$0.0} & 79.0{\scriptsize$\pm$4.0} \\
     & 4 & 86.7{\scriptsize$\pm$0.0} & 58.0{\scriptsize$\pm$1.6} & 88.6{\scriptsize$\pm$0.4} & 67.6{\scriptsize$\pm$1.0} & 96.7{\scriptsize$\pm$0.0} & 81.3{\scriptsize$\pm$3.4} \\
    \midrule
    Cns-20\% & 0.5 & \textbf{93.3{\scriptsize$\pm$0.0}} & 57.7{\scriptsize$\pm$1.5} & \textbf{90.3{\scriptsize$\pm$0.2}} & \textbf{68.6{\scriptsize$\pm$1.2}} & 96.7{\scriptsize$\pm$0.0} & 75.7{\scriptsize$\pm$1.5} \\
     & 1 & 90.0{\scriptsize$\pm$0.0} & 58.7{\scriptsize$\pm$1.6} & 90.2{\scriptsize$\pm$0.2} & 68.3{\scriptsize$\pm$1.1} & 96.7{\scriptsize$\pm$0.0} & 76.7{\scriptsize$\pm$2.6} \\
     & 2 & 86.7{\scriptsize$\pm$0.0} & \textbf{59.3{\scriptsize$\pm$1.3}} & 90.1{\scriptsize$\pm$0.3} & 68.1{\scriptsize$\pm$1.3} & 96.7{\scriptsize$\pm$0.0} & 80.0{\scriptsize$\pm$3.7} \\
     & 4 & 86.7{\scriptsize$\pm$0.0} & 57.7{\scriptsize$\pm$2.1} & 89.7{\scriptsize$\pm$0.4} & 67.6{\scriptsize$\pm$1.2} & 96.7{\scriptsize$\pm$0.0} & \textbf{82.7{\scriptsize$\pm$2.5}} \\
    \midrule
    Cns-2048 & 0.5 & 89.3{\scriptsize$\pm$1.3} & 58.0{\scriptsize$\pm$1.6} & \textbf{90.3{\scriptsize$\pm$0.3}} & 67.3{\scriptsize$\pm$1.1} & 96.7{\scriptsize$\pm$0.0} & 76.3{\scriptsize$\pm$1.0} \\
     & 1 & 89.7{\scriptsize$\pm$1.0} & 58.3{\scriptsize$\pm$1.7} & 90.1{\scriptsize$\pm$0.3} & 66.8{\scriptsize$\pm$1.1} & 96.7{\scriptsize$\pm$0.0} & 77.3{\scriptsize$\pm$2.0} \\
     & 2 & 89.7{\scriptsize$\pm$1.0} & \textbf{59.3{\scriptsize$\pm$1.3}} & 89.7{\scriptsize$\pm$0.5} & 65.9{\scriptsize$\pm$1.0} & 96.7{\scriptsize$\pm$0.0} & 80.0{\scriptsize$\pm$1.5} \\
     & 4 & 86.3{\scriptsize$\pm$1.0} & 58.0{\scriptsize$\pm$1.6} & 89.0{\scriptsize$\pm$0.6} & 65.1{\scriptsize$\pm$1.0} & 96.7{\scriptsize$\pm$0.0} & 81.7{\scriptsize$\pm$1.7} \\
    \bottomrule
  \end{tabular}

%% file: tex/app_stat.tex
\label{app:stat}
We provide the full statistical analysis supporting the trajectory difference reported in Section~\ref{sec:investigation}. Both analyses are computed on the existing completion pools, requiring no additional generation.

\textbf{AUROC.} Table~\ref{tab:auroc} reports the area under the ROC curve for separating correct from incorrect completions on hard problems (Pass@1 $<20\%$), using either mean confidence or the consilience score as the ranking signal. Mean confidence is at chance (0.47--0.50), while consilience is a reliable discriminator (0.61--0.62); the 95\% confidence intervals are non-overlapping, confirming that the separation is not attributable to sampling noise.

\begin{table}[h]
  \centering
  \caption{AUROC (95\% CI) for separating correct from incorrect completions on hard problems (Pass@1 $<20\%$).}
  \input{tab/stat_sig}
  \label{tab:auroc}
\end{table}

\textbf{Paired test of the trajectory rise.} The $\pm$SEM band in Figure~\ref{fig:conf-diff} reflects \emph{between}-problem variance, whereas the initial$\rightarrow$final rise is a \emph{within}-problem effect. We therefore test the per-problem quantity $\Delta = C_{final} - C_{initial}$ (last-20\% minus first-20\%, the same window used by our selection method) with a paired test across hard problems, reported in Table~\ref{tab:paired}. The rise is significant on both LiveCodeBench panels (paired $t$: $p=0.011, 0.018$; Wilcoxon $p=0.011, 0.018$; Cohen's $d_z = 0.37, 0.35$) and positive on GPQA (Wilcoxon $p=0.042$). The HMMT hard subset ($n=7$) is underpowered for a significance claim. Because the 20\% window is fixed by the method design, this test is pre-specified rather than chosen post-hoc to maximize significance.

\begin{table}[h]
  \centering
  \caption{Paired test of the per-problem rise $\Delta=C_{final}-C_{initial}$ (last-20\% minus first-20\%) on hard problems.}
  \input{tab/paired_test}
  \label{tab:paired}
\end{table}

%% file: tab/stat_sig.tex
\begin{tabular}{lcc}
    \toprule
    \textbf{Metric} & \textbf{Qwen} & \textbf{GPT-OSS-120B} \\
    \midrule
    Mean confidence & 0.50 {\scriptsize[0.48, 0.52]} & 0.47 {\scriptsize[0.44, 0.50]} \\
    Consilience & \textbf{0.61 {\scriptsize[0.59, 0.64]}} & \textbf{0.62 {\scriptsize[0.59, 0.65]}} \\
    \bottomrule
\end{tabular}

%% file: tab/paired_test.tex
\begin{tabular}{lccccc}
    \toprule
    \textbf{Panel} & \textbf{$n$} & \textbf{mean $\Delta$} & \textbf{95\% CI} & \textbf{paired $t$ ($p$)} & \textbf{Cohen's $d_z$} \\
    \midrule
    LCB / GPT-OSS-120B & 37 & $+0.077$ & $[+0.007, +0.148]$ & $p=0.011$ & $+0.37$ \\
    LCB / Qwen        & 56 & $+0.077$ & $[+0.018, +0.136]$ & $p=0.018$ & $+0.35$ \\
    GPQA / Qwen       & 72 & $+0.036$ & $[-0.002, +0.073]$ & $p=0.063$ & $+0.22$ \\
    HMMT / GPT-OSS-20B & 7 & $+0.030$ & $[-0.067, +0.127]$ & $p=0.48$  & $+0.28$ \\
    \bottomrule
\end{tabular}

%% file: tex/app_nq.tex
\label{app:nq}
To evaluate consilience at a small, practical sample size, under extended baselines, and on a data distribution outside the mathematics/code domains of the main paper, we conduct an additional experiment on Natural Questions~\citep{kwiatkowski2019natural}, a common-sense open-domain question answering benchmark. We use a 2K-question subsample and GPT-OSS-20B, with a budget of $n=8$ generations. To ensure a fair comparison, every method is granted the same generation budget of eight total passes, allocated according to its design: consilience and DeepConf select from 8 parallel samples; Self-Refine~\citep{madaan2023selfrefineiterativerefinementselffeedback} runs 8 sequential refinement iterations; and Self-Verify~\citep{saunders2022selfcritiquingmodelsassistinghuman} runs 4 parallel threads each with one additional self-verification round.

\begin{table}[h]
    \centering
    \caption{Natural Questions (2K subsample, GPT-OSS-20B) at a practical budget of $n=8$. All methods use eight total generation passes. Wall-time is measured on identical hardware.}
    \input{tab/nq}
    \label{tab:nq}
\end{table}

Table~\ref{tab:nq} reports the results. At the practical $n=8$ budget, consilience improves over Pass@1 by $+4.2$ points and is competitive with the strongest baseline, Self-Verify ($30.1$ vs.\ $30.4$), while being fully parallelizable—it completes in $36$ minutes versus $59$ minutes for Self-Verify and $101$ minutes for the sequential Self-Refine. Self-Refine does not help on this dataset. Two points are worth noting. First, gains persist at a small budget on a distribution distinct from math and code, indicating that the consilience signal is not specific to the formal-reasoning domains of the main experiments. Second, judge- and refinement-based methods incur substantial wall-time because they cannot be fully parallelized, and they scale poorly for long free-form outputs (e.g., LiveCodeBench solutions), where feeding all candidates into a single verification call can quickly exhaust the context window. Consilience avoids this by scoring each completion independently from its own logits. We also note that consilience is complementary to judge-based selection: a judge could additionally assess the exploratory diversity of the reasoning process rather than only final-answer consensus, which we leave as future work.

%% file: tab/nq.tex
\begin{tabular}{lcc}
    \toprule
    \textbf{Method} & \textbf{Accuracy (\%)} & \textbf{Wall-time} \\
    \midrule
    Pass@1 & 25.9 & 14 min \\
    DeepConf ($n=8$) & 26.7 & 37 min \\
    Consilience ($n=8$) & \textbf{30.1} & 36 min \\
    Self-Refine (8 iter) & 26.0 & 101 min \\
    Self-Verify ($4\times2$) & 30.4 & 59 min \\
    \bottomrule
\end{tabular}

%% file: tex/app_more_rel.tex
\label{app:more_rel}
\textbf{Verifier-based and search-based TTS.} Beyond the confidence-based methods discussed in the main text, a large body of parallel TTS work relies on external signals. Programmatic verifiers such as compilers~\citep{zheng2025opencodeinterpreterintegratingcodegeneration,le2022coderlmasteringcodegeneration} and formal proof checkers provide ground-truth feedback, and search algorithms combine sequential and parallel rollouts to select promising partial traces~\citep{yao2023treethoughtsdeliberateproblem,wang2025rolloutcountsoptimalresource}. On the aggregation side, Self-Consistency with early stopping~\citep{li2024escapeskyhighcostearlystopping}, Process Reward Models that verify intermediate steps~\citep{wang-etal-2024-math}, recursive/ensemble self-aggregation~\citep{venkatraman2026recursiveselfaggregationunlocksdeep,kim2026scalingtesttimecomputeagentic}, and self-verification~\citep{saunders2022selfcritiquingmodelsassistinghuman} improve robustness at the cost of extractable answers, trained verifiers, or substantial aggregation overhead. Consilience instead requires only the model's own logits and applies to free-form generation.

\textbf{Using Entropy and Confidence in LLMs.} In proximal area of output caliberation, confidence and entropy are also widely acknowledged as an important signal to determine output quality~\citep{geng-etal-2024-survey,fadeeva2024factcheckingoutputlargelanguage,duan2024shiftingattentionrelevancepredictive}. Entropy use is also adopted in reinforcement learning research, as this signal frequently serve as effective RL optimization targets. Leveraging this, works like Test-Time Reinforcement Learning~\citep{zuo2025ttrltesttimereinforcementlearning} utilize internal metrics to lift the reliance on external verifiers, typically minimizing sequence entropy~\citep{agarwal2025unreasonableeffectivenessentropyminimization} to drive high-confidence consensus. Within Reinforcement Learning with Verifiable Rewards, on the other hand, researchers utilize entropy maximization as a proactive global training perturbation; frameworks such as DACE~\citep{li2025knowexploredifficultyawarecertainty}, UCAS~\citep{xie2026unlockingexplorationrlvruncertaintyaware}, and Reasoning with Exploration~\citep{cheng2025reasoningexplorationentropyperspective} explicitly integrate entropy bonuses to penalize convergence and force diverse search paths. \citet{zeng2026pruningunsurprisingefficientllm} uses entropy to quantify surprise in LLM generation to effectively curate data. Our consilience framework connects to these efforts by utilizing the same internal probability distributions, but diverges significantly in both application and conceptualization. Rather than acting as a global training perturbation that steers intermediate training procedure for exploration, consilience serves as a zero-shot inference signature leading to robust reasoning process. Moreover, we demonstrate that the value of exploration is phase-dependent: high entropy is profoundly beneficial during initial hypothesis branching, but uniformly high entropy throughout a sequence signifies total confusion rather than robust exploration.

%% file: tex/app_qualitative.tex
\label{app:qualitative}
\begin{figure}[t]
    \centering
    \input{figures/fig_case_study}
    \caption{Confidence in first 2048 tokens for two completions. Consilience pick, despite lower initial confidence, spent more time on critical design choice, and lead to final correct in this particularly hard problem.}
    \label{fig:case_study}
\end{figure}
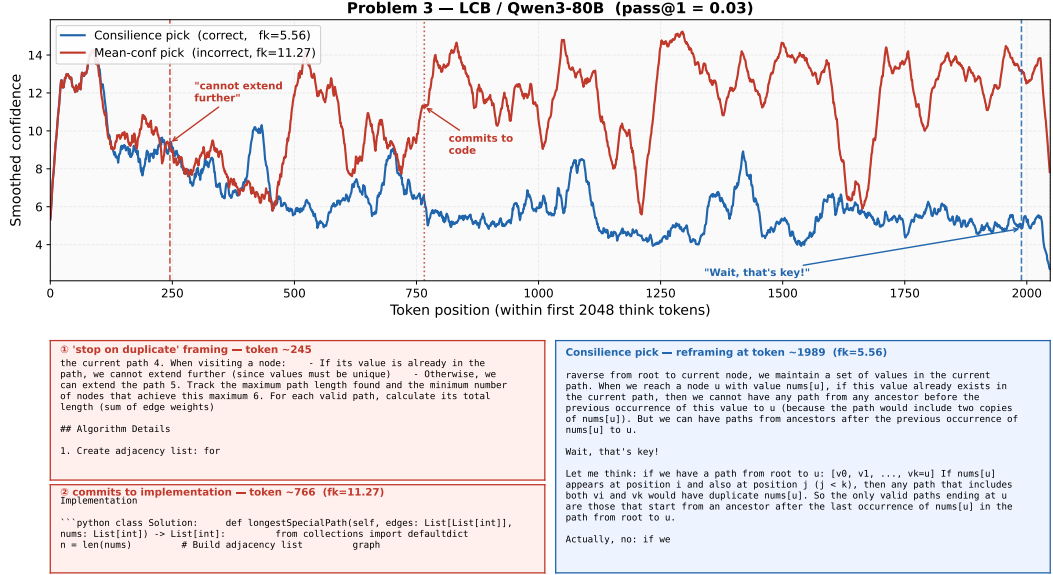
\vspace{4pt}

Figure~\ref{fig:case_study} traces the per-token confidence of two completions
produced by \textsc{Qwen3-80B} on the problem below (pass@1\,=\,0.03; only 4 of
128 completions are correct).
The \emph{consilience pick} ($f_k{=}5.56$, blue) is correct; the
\emph{mean-confidence pick} ($f_k{=}11.27$, red) is not.

\begin{tcolorbox}[colback=blue!5!white, colframe=blue!60!black, fontupper=\small\ttfamily,
                  title={\small\textbf{Problem (LCB \#3)}}, left=4pt, right=4pt,
                  top=3pt, bottom=3pt]
Given an undirected tree rooted at node~0 with $n$ nodes, where each node has a
value \texttt{nums[i]} and each edge has an integer length.
A \emph{special path} is a downward path (ancestor $\to$ descendant) in which all
node values are distinct.
Return \texttt{[max\_length, min\_nodes]}: the length of the longest special path,
and the minimum number of nodes among all such longest paths.
($n \le 5{\times}10^4$, \texttt{nums[i]}$\le 5{\times}10^4$.)
\end{tcolorbox}

Both completions open with the same sketch: DFS from the root, maintain a set of
seen values, stop when a duplicate is encountered.
The mean-confidence completion accepts this framing immediately (marker 1):
\emph{``if its value is already in the path, we cannot extend further''}---and
never revisits it.
This \emph{stop-on-duplicate} assumption translates directly into a
\texttt{return} statement in the final code, silently pruning entire subtrees:
once a node with a repeated value is reached, none of its descendants are
explored or scored.
The bug is fatal on any input where a long valid path passes through a node
whose value already appeared higher in the tree, because the stop prevents the
DFS from ever discovering that valid path.
Confident in its framing, the model commits to writing code at marker 2.

The consilience completion, by contrast, spends the entire first 2{,}048 tokens
questioning the same framing.
Recognising that the path may start at \emph{any} ancestor---not just the
root---it repeatedly asks: which ancestors are actually valid starting points for
a path ending at the current node?
At marker 3 it finds the answer: \emph{``we cannot have any path from any
ancestor \textbf{before} the previous occurrence of \texttt{nums[u]} to~$u$
\ldots\ but we can have paths from ancestors \textbf{after} the previous
occurrence \ldots\ So the only valid paths ending at~$u$ are those that start
from an ancestor after the last occurrence of \texttt{nums[u]}.\ Wait, that's
key!''}
The model continues to reason for thousands of
tokens before arriving at the concrete algorithm for this problem. It identifies the right
question to ask, a question the over-confident completion never posed because it
had already committed to code.
 

%% file: figures/fig_case_study.tex
\tikzset{
  cs badge/.style={circle, fill=#1, text=white, font=\fontsize{5.5}{6}\selectfont\bfseries, inner sep=0.7pt, minimum size=7.5pt},
  cs quote/.style={draw=#1, line width=0.6pt, fill=#1!6, rounded corners=1.5pt, inner sep=4pt, align=left,
    font=\ttfamily\fontsize{5.6}{6.9}\selectfont, text=csInk},
  cs qhead/.style={font=\scriptsize\bfseries, text=csInk},
}
\begin{tikzpicture}
\begin{axis}[
  name=plot,
  cs panel, cs legend,
  width=12.75cm, height=3.3cm,
  xmin=0, xmax=2048, ymin=2, ymax=16.2,
  xtick={0,250,500,750,1000,1250,1500,1750,2000},
  ytick={4,6,8,10,12,14},
  xlabel={Token position (first 2048 reasoning tokens)},
  ylabel={Smoothed confidence},
  title={Problem 3, LCB / Qwen3-80B (pass@1 = 0.03)},
  legend style={at={(axis description cs:0.5,1.07)}, anchor=south},
  title style={yshift=13pt},
]
\addplot[cs line, csCons, line width=0.75pt] table[x=token, y=cons] {figures/data/case_trace.dat};
\addlegendentry{Consilience pick (correct, $f_k$\,=\,5.56)}
\addplot[cs line, csOver, line width=0.75pt] table[x=token, y=mean] {figures/data/case_trace.dat};
\addlegendentry{Mean-confidence pick (incorrect, $f_k$\,=\,11.27)}
\draw[csOver, line width=0.6pt, densely dashed] (axis cs:245,2) -- (axis cs:245,16.2);
\draw[csOver, line width=0.6pt, densely dotted] (axis cs:766,2) -- (axis cs:766,16.2);
\draw[csCons, line width=0.6pt, densely dashed] (axis cs:1989,2) -- (axis cs:1989,16.2);
\node[cs badge=csOver] at (axis cs:245,16.2) {1};
\node[cs badge=csOver] at (axis cs:766,16.2) {2};
\node[cs badge=csCons] at (axis cs:1989,16.2) {3};
\node[cs note, anchor=west, fill=white, fill opacity=0.85, text opacity=1] at (axis cs:262,12.6) {``cannot extend further''};
\node[cs note, anchor=west, fill=white, fill opacity=0.85, text opacity=1] at (axis cs:783,9.6) {commits to code};
\node[cs note, anchor=east, fill=white, fill opacity=0.85, text opacity=1] at (axis cs:1972,3.2) {``Wait, that's key!''};
\end{axis}

\node[cs quote=csOver, anchor=north west, text width=6.42cm] (q1) at ([yshift=-0.25cm]plot.outer south west) {%
  {\normalfont\tikz[baseline=-0.6ex]\node[cs badge=csOver]{1};\ \textbf{\scriptsize `Stop on duplicate' framing, token $\sim$245}}\\[2pt]
  the current path 4. When visiting a node: - If its value is already in the path, we cannot extend further (since values must be unique) - Otherwise, we can extend the path 5. Track the maximum path length found and the minimum number of nodes that achieve this maximum 6. For each valid path, calculate its total length (sum of edge weights)\\[2pt]
  \#\# Algorithm Details\\[1pt]
  1. Create adjacency list: for};
\node[cs quote=csOver, anchor=north west, text width=6.42cm] (q2) at ([yshift=-0.18cm]q1.south west) {%
  {\normalfont\tikz[baseline=-0.6ex]\node[cs badge=csOver]{2};\ \textbf{\scriptsize Commits to implementation, token $\sim$766 ($f_k$\,=\,11.27)}}\\[2pt]
  Implementation\\[1pt]
  \`{}\`{}\`{}python class Solution: def longestSpecialPath(self, edges: List[List[int]], nums: List[int]) -> List[int]: from collections import defaultdict n = len(nums) \# Build adjacency list graph};
\node[cs quote=csCons, anchor=north east, text width=6.42cm] (q3) at ([yshift=-0.25cm]plot.outer south west -| plot.south east) {%
  {\normalfont\tikz[baseline=-0.6ex]\node[cs badge=csCons]{3};\ \textbf{\scriptsize Consilience pick: reframing at token $\sim$1989 ($f_k$\,=\,5.56)}}\\[2pt]
  \ldots traverse from root to current node, we maintain a set of values in the current path. When we reach a node u with value nums[u], if this value already exists in the current path, then we cannot have any path from any ancestor before the previous occurrence of this value to u (because the path would include two copies of nums[u]). But we can have paths from ancestors after the previous occurrence of nums[u] to u.\\[2pt]
  Wait, that's key!\\[2pt]
  Let me think: if we have a path from root to u: [v0, v1, \ldots, vk=u] If nums[u] appears at position i and also at position j (j < k), then any path that includes both vi and vk would have duplicate nums[u]. So the only valid paths ending at u are those that start from an ancestor after the last occurrence of nums[u] in the path from root to u.\\[2pt]
  Actually, no: if we};
\end{tikzpicture}

%% file: tex/app_prompts.tex
\label{app:prompts}

For all evaluations, we utilized the minimal and default prompt templates supplied by their respective benchmarks. We introduced only slight modifications to the instructions to enforce strict structural formatting (such as \texttt{\textbackslash boxed\{\}} or specific Markdown delimiters) to ensure reliable, automated answer extraction. The exact system and user prompts for each dataset are detailed below.

\begin{tcolorbox}[colback=blue!5!white, colframe=blue!60!black, title=LiveCodeBench (LCB) Prompt]
\textbf{System Message:}\\
You are an expert Python programmer. Your task is to solve a coding problem by first creating a plan and then writing clean, efficient code.

**IMPORTANT INSTRUCTIONS:**
1. **Show Thinking** In your response, first describe your detailed step-by-step thinking process, how you understand the problem, and how you arrive at your solution. You then provide the implementation plan of your code.
2. **Clean Code Block:** After the plan, provide the complete and final code solution in a single, clean markdown block.
3. **NO THINKING IN CODE:** The code block MUST NOT contain any comments that explain your thought process or step-by-step logic.

\vspace{2mm}
\textbf{User Message (Without Starter Code):}\\
You will be given a question (problem specification) and will generate a correct Python program that matches the specification and passes all tests beyond provided examples in the most optimal complexity.

Question:
\{input\_text\}

First, provide your plan. Then, write code to read the inputs from stdin, solve the problem, and write the answer to stdout (do not directly test on the sample inputs). Enclose your code within a single ```python block.

```python \# YOUR CODE HERE ```
\end{tcolorbox}

For Qwen model, despite explicit instruction to contraint its thinking outside the code block, it still tends to put big trunk of thinking process as comments in the final code.

\begin{tcolorbox}[colback=blue!5!white, colframe=blue!60!black, title=AIME Prompt]
\textbf{System Message:}\\
You are an expert mathematician solving AIME (American Invitational Mathematics Examination) problems. Provide clear step-by-step reasoning and clearly mark your final answer using \textbackslash boxed\{\} notation.

\vspace{2mm}
\textbf{User Message:}\\
\{question\}

Please reason step by step, and put your final answer within \textbackslash boxed\{\}.
\end{tcolorbox}

\begin{tcolorbox}[colback=blue!5!white, colframe=blue!60!black, title=HMMT Prompt]
\textbf{System Message:}\\
You are a helpful assistant.

\vspace{2mm}
\textbf{User Message:}\\
Please reason step by step, and put your final answer within \textbackslash boxed\{\}.

\{problem\}
\end{tcolorbox}

\begin{tcolorbox}[colback=blue!5!white, colframe=blue!60!black, title=GPQA Prompt]
\textbf{System Message:}\\
You are an expert at answering graduate-level science questions. Answer the following multiple-choice question by analyzing each option carefully.

\vspace{2mm}
\textbf{User Message:}\\
\{Question\}

(A) \{A\}\\
(B) \{B\}\\
(C) \{C\}\\
(D) \{D\}

Express your final answer as the corresponding option 'A', 'B', 'C', or 'D'. Put your answer within \textbackslash boxed\{\}.
\end{tcolorbox}

%% file: tex/app_agentic.tex
\label{app:agentic}
In multi-step, interactive agentic environments such as SWE-bench, applying test-time scaling (TTS) and consilience selection to every generation step is both computationally inefficient and theoretically unnecessary. Many steps in an agentic trajectory are rudimentary information-gathering actions—such as listing directories or reading file chunks—that do not necessitate deep cognitive search or complex reasoning. Therefore, we constrain the application of our metric strictly to critical editing steps: the junctures where the model actually formulates the logic to modify the codebase.

To integrate consilience into a sequential workflow without derailing the agent's overarching strategy, we design a targeted interception harness. The execution proceeds as follows:
\begin{enumerate}
    \item \textbf{Base Sampling and Detection:} For any given step, we initially sample a single standard completion. We analyze this response via keyword matching to determine if it constitutes a file-editing action (specifically checking for: \texttt{sed -i}, \texttt{cat <<}, \texttt{tee }, \texttt{> /}, \texttt{patch }, or \texttt{EOF}).
    \item \textbf{Conditional Triggering:} If the step is merely exploratory (no keywords detected), it is executed normally. If an editing keyword is present, and the bash command is larger then L lines, the step is flagged as a critical reasoning node. (We use $L=40$ for GPT-OSS-120B, and $L=100$ for Qwen3-Coder-Next).
    \item \textbf{Parallel Generation and Filtering:} At a flagged node, we sample $K$ parallel completions. To ensure the agent adheres to the established workflow and intent of the base trajectory, we filter these $K$ candidates, retaining only those that utilize the same editing keywords identified in the initial step.
    \item \textbf{Consilience Selection:} Finally, we evaluate the isolated reasoning phases of the filtered pool and apply our consilience metric to select the most structurally robust editing command for execution.\end{enumerate}

We note that this keyword-triggered interception is an intentionally coarse harness. Its primary function is to serve as a minimal viable integration to demonstrate that consilience effectively identifies high-quality reasoning in open-ended, sequential agentic tasks where exact-match voting is impossible. Developing more sophisticated, dynamic TTS triggering mechanisms for autonomous agents remains a compelling avenue for future work.